%% file: author.tex
\documentclass{article}
\usepackage{jfrExamplee}
\usepackage{graphicx}

\usepackage{mathptmx}       
\usepackage{helvet}         
\usepackage{courier}        
\usepackage{type1cm}        
\usepackage{subcaption}
\usepackage{gensymb}
\usepackage{textcomp} 
\usepackage[section]{placeins}
\usepackage{url}

\usepackage{threeparttable}
\usepackage{siunitx} 
\usepackage{makeidx}         
\usepackage{graphicx}        
\usepackage{multicol}        
\usepackage[bottom]{footmisc}
\usepackage[colorinlistoftodos]{todonotes}
\usepackage{units}
\usepackage{soul} 
\usepackage{enumitem}

\usepackage{tikz} 

\usepackage{hyperref}

\usepackage{multirow}
\usepackage{bigstrut}
\usepackage{booktabs}
\usepackage{pbox} 
\usepackage{makecell}
\usepackage{array}
\usepackage{tablefootnote}          
\usepackage{scrextend}              

\usepackage{calrsfs}
\DeclareMathAlphabet{\pazocal}{OMS}{zplm}{m}{n}

\usepackage{multicol} 
\usepackage{tablefootnote}

\newcommand{\mytilde}{\raise.17ex\hbox{$\scriptstyle\mathtt{\sim}$}}
\newcommand{\myless}{\raise.27ex\hbox{\scriptsize\textless $\scriptstyle\mathtt{\sim}$}}
\newcommand{\mymore}{\raise.27ex\hbox{\scriptsize\textgreater $\scriptstyle\mathtt{\sim}$}}

\newcommand{\pie}[1]{%
\lower.5ex\hbox{\begin{tikzpicture}
 \draw (0,0) circle (1ex);\fill (1ex,0) arc (0:#1:1ex) -- (0,0) -- cycle;
\end{tikzpicture}}%
}

\usepackage{amsmath,amssymb}
\usepackage{bm}

\usepackage{gensymb}

\title{Traversing Steep and Granular Martian Analog Slopes With a Dynamic Quadrupedal Robot}

\author{
  \begin{minipage}{35em}
    \begin{center}
      Hendrik Kolvenbach\textsuperscript{1}, Philip Arm\textsuperscript{1}, Elias Hampp \textsuperscript{1}, Alexander Dietsche\textsuperscript{1}, Valentin Bickel\textsuperscript{2,3}, Benjamin Sun\textsuperscript{1}, Christoph Meyer\textsuperscript{1}, Marco Hutter\textsuperscript{1}\\
    \end{center}
  \end{minipage} \\
  \\
  \textsuperscript{1} Robotic Systems Lab\\
  ETH Zurich, Switzerland \\
  \And \\
  \textsuperscript{2} Department of Earth Sciences\\
  ETH Zurich, Switzerland  \\
    \And \\
  \textsuperscript{3} Department Planets and Comets\\
  Max Planck Institute for Solar System Research, Germany  \\
}

\begin{document}
\maketitle


\begin{abstract}
Celestial bodies such as the Moon and Mars are mainly covered by loose, granular soil, a notoriously challenging terrain to traverse with (wheeled) robotic systems. Here, we present experimental work on traversing steep, granular slopes with the dynamically walking quadrupedal robot SpaceBok. To adapt to the challenging environment, we developed passive-adaptive planar feet and optimized grouser pads to reduce sinkage and increase traction on planar and inclined granular soil. Single-foot experiments revealed that a large surface area of \SI{110}{cm^2} per foot reduces sinkage to an acceptable level even on highly collapsible soil (ES-1). Implementing several \SI{12}{mm} grouser blades increases traction by 22\% to 66\% on granular media compared to grouser-less designs. Together with a terrain-adapting walking controller, we validate - for the first time - static and dynamic locomotion on Mars analog slopes of up to 25\degree (the maximum of the testbed). We evaluated the performance between point- and planar feet and static and dynamic gaits regarding stability (safety), velocity, and energy consumption. We show that dynamic gaits are energetically more efficient than static gaits but are riskier on steep slopes. Our tests also revealed that planar feet's energy consumption drastically increases when the slope inclination approaches the soil's angle of internal friction due to shearing. Point feet are less affected by slippage due to their excessive sinkage, but in turn, are prone to instabilities and tripping. We present and discuss safe and energy-efficient global path-planning strategies for accessing steep topography on Mars based on our findings.

\end{abstract}

\section{Introduction}\label{sec:Introduction}
\subsection{Motivation}\label{sec:leggedintroduction}



Regolith is the predominant medium covering the surface of Mars, the Moon, and many other celestial bodies. Its often loose, granular, and heterogeneous nature complicates the traversal with wheeled robots. Several exploration rovers encountered highly unfavorable soil conditions, leading to excessive sinkage and wheel slip \cite{gonzalez2018slipjfr}. Those, in some cases critical events, have severely impacted the timelines of missions. For example, it took five weeks to free the \textit{Opportunity} rover from loose sand in 2006 \cite{young2006opportunitystuck}, and rover trajectories are frequently adjusted to avoid challenging terrain \cite{arvidson2016megaripple}. The potentially worst situation occurred in 2009 when the \textit{Spirit} rover got stuck in an aeolian sand deposit and was unable to recover, which ultimately terminated the mission \cite{webster2009nasa}. The risks have restricted the systems to explore only moderately sloped areas, although interesting targets for future missions require the robot to overcome steep environments (more than $>$15°) such as craters \cite{seeni2010, potts2015, steenstra2016, Czaplinski2021schroedinger}. Future missions are also looking into improved and accelerated mobility to traverse longer distances while operating under tight time constraints \cite{rodriguez2019highspeed}.

Legged robots have significant potential for fast and safe locomotion in complex or partially unknown environments. The high maneuverability due to the multi-degree-of-freedom (DOF) legs allows for precise foot placement to overcome obstacles and control of the foot contact force to avoid slip. Legged locomotion allows the use of various gait patterns, which can be altered as a function of gravity, encountered terrain, and the desired speed, thus reaching a high mobility and efficacy level. For example, a safe, static gait could be used in steep or unknown terrain, while a more efficient and faster dynamic gait could be used on known, flat terrain to maximize speed and minimize energy consumption.

So far, only a few, mostly multilegged, walking robots have been developed with the intention of usage in space \cite{kolvenbach2017iac}. Examples include the eight-legged \textit{Dante} project at Carnegie Mellon in 1993 \cite{wettergreen1993dante1, Bares1999danteII}, the six-legged walking robots started at DFKI around 2007, \textit{Scorpion} \cite{dirk2007bio}, \textit{SpaceClimber} \cite{bartsch2012spaceclimber} and FZI's \textit{Lauron V} \cite{roennau2014iros} in 2014.  Our group proposed the use of a quadrupedal robot for ESA's Lunar Robotics Challenge in 2010, during which the robot performed an inching motion to crawl on the terrain \cite{remy2010alof}. Scaled plates on the bottom of the robot and the shanks helped to generate traction but lead to significant side-slip on the slope. In conclusion, the systems mentioned above can walk robustly on uneven and sloped areas by using static gaits, whereby a minimum of three feet in ground contact ensures a statically stable stance at all times. However, being reduced to static gaits and inching motions limits the achievable traversal speed, agility, and energy efficiency. 

Another group of limbed exploration robots is formed by leg-wheel hybrids, where wheels are attached to an actuated linkage. In the context of space exploration, this concept has been demonstrated by DFKI's \textit{SHERPA} \cite{cordes2014active} with 24 DOF,  NASA's \textit{ATHLETE} with 36 DOF \cite{Wilcox2007athlete} or the latest iteration of \textit{RoboSimian} with 32 DOF \cite{reid2019fsr}. While the multimodal systems show promising performance in flat terrain and high mobility in unstructured environments, the high number of articulated joints drastically increases the system complexity, potential points of failure, number of entry points for dust, and overall mass. 

\begin{figure}[!tb]
\centering
   \includegraphics[scale=0.53]{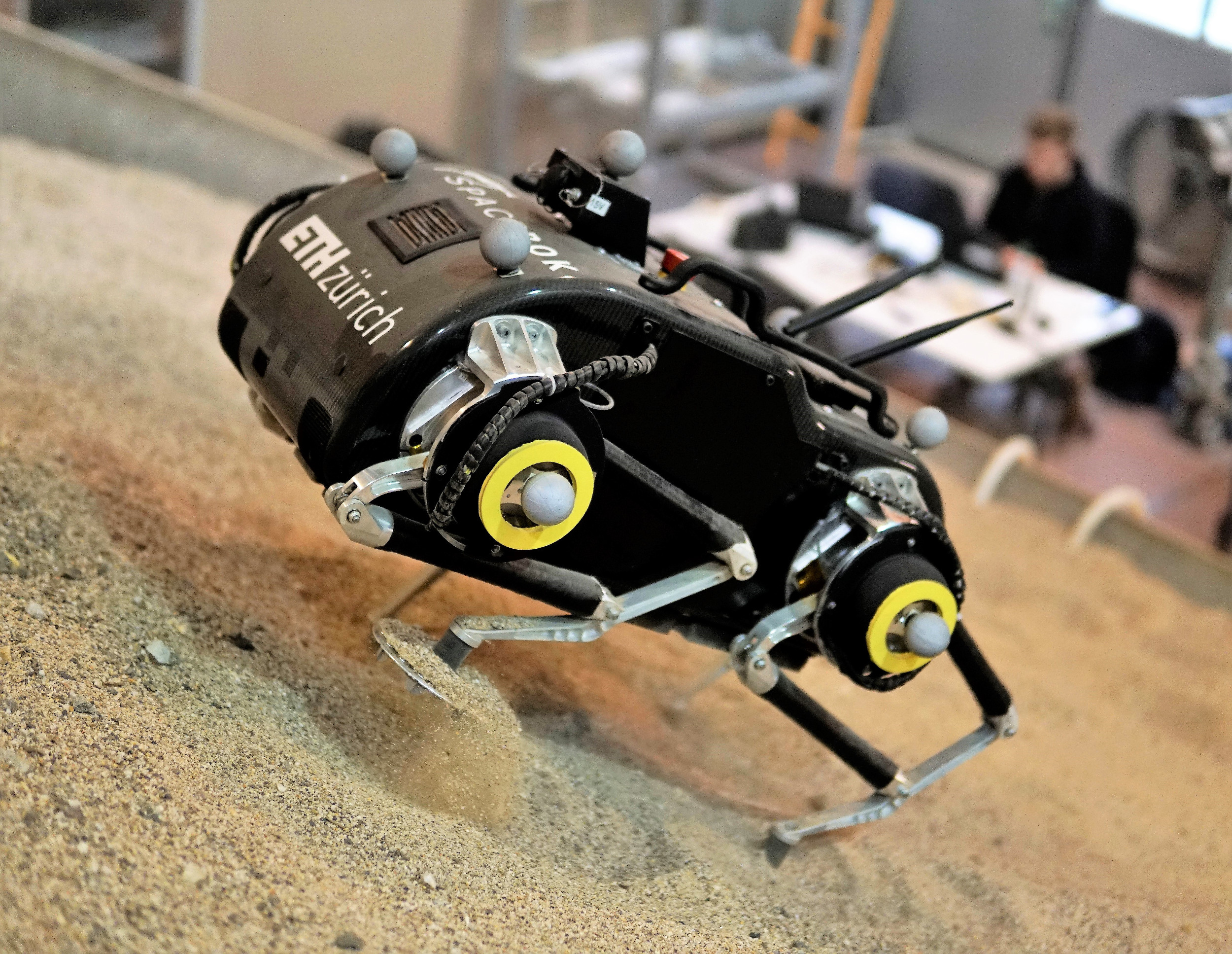}
    \caption{\textit{SpaceBok}, equipped with adaptive planar feet, walking up a \unit[25]{\degree} inclined Mars analog slope. }
    \label{Fig:spacebok_title}
\end{figure}

Meanwhile, general-purpose legged robots for terrestrial use, such as \textit{ANYmal} \cite{hutter2017anymal}, MIT \textit{Cheetah} \cite{bledt2018cheetah3}, MIT \textit{Mini Cheetah} \cite{katz2019minicheetah}, \textit{HYQ2Max} \cite{semini2017hyq2max}, \textit{Spot} from Boston Dynamics \cite{spotmini}, or the robots from Unitree Robotics, \textit{Laikago} \cite{laikago} and \textit{Aliengo} \cite{aliengo} have made remarkable progress over the last decade. It has been shown that this group of dynamic legged robots can walk quickly over rough terrain, recover from falls, navigate unstructured environments, and operate for several hours on a single battery charge. The increasing maturity of the technology has allowed the robots to leave the lab and start being used in real-world applications \cite{bellicoso2018applications, kolvenbach2020jfr}. In this context, our group and the German Aerospace Center have started investigating dynamic legged locomotion for planetary exploration with the quadrupedal robots \textit{SpaceBok} \cite{arm2019icra} (Figure \ref{Fig:spacebok_title}) and \textit{Bert} \cite{lakatos2018bert}, respectively.

However, it remains an open question how a dynamically walking legged robot can cope with unfavorable, granular, and sloped environments found on celestial bodies like Mars. Thus, the contribution of this paper is threefold:
\begin{enumerate}[topsep=0pt,itemsep=-1ex,partopsep=1ex,parsep=1ex]
  \item We investigate sand-walking feet to enable safe and efficient traversal of granular soils.
  \item We validate and evaluate static and dynamic walking gaits on martian analog slopes of up to 25\degree.
  \item We discuss global path-planning strategies for accessing craters on Mars while taking safety and energy constraints into account.
\end{enumerate}

With this work, we want to showcase the potential of dynamically walking legged robots for future planetary exploration, and help to guide future research. 

This paper is structured as follows: First, we describe the \textit{SpaceBok} robot with a focus on locomotion control in Sec.~\ref{sec:spacebok_description}. Next, the development and design of specialized walking feet and respective testing is presented in Sec.~\ref{sec:spacebok_feet_design}. We performed slope walking experiments at the ESA \textit{ExoMars} rover locomotion testbed and report our findings in Sec.~\ref{sec:ruag_tests}. The energy consumption, which we derived from the test campaign, was used in conjunction with a global path planner to find energy-efficient journeys on Mars, as presented in Sec.~\ref{sec:mars_path_strategies}. Next, our results are discussed in Sec.~\ref{sec:slope_climbing_discussion} before drawing a conclusion in Sec.~\ref{sec:slope_climbing_conclusion}. 


\section{SpaceBok}\label{sec:spacebok_description}
\subsection{Hardware design}\label{sec:spacebok}
We developed the first version of the quadrupedal robot \textit{SpaceBok} in 2018 to research dynamic legged locomotion for planetary exploration. The robot is not space-graded but serves as a technology demonstration platform for various surface mobility experiments and is updated frequently to new experiments' needs.

\textit{SpaceBok's} development was governed by a focus on lightweight design, a leg structure that allows the actuators to work in conjunction, and a control structure that allows for static and dynamic walking gaits. Each leg has two actuated Degrees of Freedom (DOF), which allow for hip flexion/extension and knee flexion/extension. The two actuators are placed co-axially and power the leg's parallel mechanism (Figure \ref{fig:leg_kinematics}). 

We omitted the hip abduction/adduction DOF to save weight and decrease complexity, which will be key to decrease the costs of a potential future space mission. The robot's hip height is 500mm, and the platform weighs \SI{22}{kg} in the current configuration. The system is equipped with a small-scale Intel i7 computer (Intel NUC) that executes the control software. A high-performance Inertial Measurement Unit (IMU) (Vectornav VN100) provides pose estimates, and a lithium-polymer battery (Swaytronic 12S 6000 mAh) with an in-house developed Battery Management System (BMS) powers the robot. The BMS allows monitoring of energy consumption and charging of the battery in case energy is recuperated. A 20mF capacitor is placed between the battery and the consumers to smooth the readings.

We published a detailed hardware overview of the platform in earlier work \cite{arm2019icra}. For this work, we added temperature sensors as a safety feature to the actuators and sealed the robot to reduce the amount of dust ingress into the system. We also developed special feet to allow traversal of sandy terrain, as discussed in Sec. \ref{sec:spacebok_feet_design}.

\subsection{Gait selection}\label{sec:gaitselection}
Compared to other prototypes of legged robots for space exploration, \textit{SpaceBok} employs not only static but also dynamic walking gaits. A static walking gait is characterized by always keeping three feet in ground contact. This allows the robot to stand stable but limits the achievable velocity since only one leg can be moved at a time.

In nature, almost only dynamic gaits such as trotting and galloping are observed. During dynamic walking or running, there are moments when less than three feet are in ground contact, ultimately allowing for faster speed at the expense of increased control effort.

\begin{figure}[!tb]
\centering
   \begin{subfigure}[]{0.5\textwidth}
   \centering
   \input{Figures/control/03_staticwalk.tikz}
    \caption{Static Walk}
    \end{subfigure}%
    \begin{subfigure}[]{0.5\textwidth}
    \centering
   \input{Figures/control/04_trot.tikz}
    \caption{Trot}
    \end{subfigure}%
    \caption{Gait pattern illustration of the gaits used for slope climbing. The acronyms on the left indicate the respective leg (R = Right, L = Left, F = Front, H = Hind). TO = Take-off, TD = Touch-down. The static walking gait has a phase shift of 25\%, the trot a phase shift of 50\%.}
    \label{fig:gaitpattern}
\end{figure}
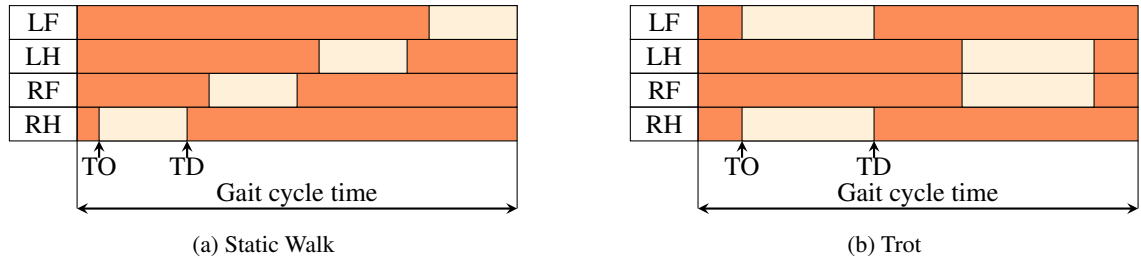

The gaits used in this study are designed according to common quadrupedal gaits described by Hildebrand \cite{Hildebrand1989gaits}. We chose a static- and a trotting gait, as illustrated in Figure \ref{fig:gaitpattern}. The specific design parameters include the total gait cycle time in seconds, the duty factor, i.e., the percentage of the full gait cycle that a specific leg is in ground contact, and the foot's step height during the swing phase. We express the take-off and touch-down times as a ratio of the total gait phase. The exact timings can be found in Sec. \ref{sec:ruag_results_performance}.

\subsection{Locomotion control}\label{sec:control}
An essential factor of the robot's ability to climb sandy slopes is the locomotion controller. For this study, we extended \textit{SpaceBok's} locomotion controller from previous work by the ability to estimate and adapt to the slope of the terrain. 

\subsubsection{Terrain estimation}\label{sec:terrainestimation}
\begin{figure}[!tb]
\centering
   \includegraphics[width=0.7\textwidth]{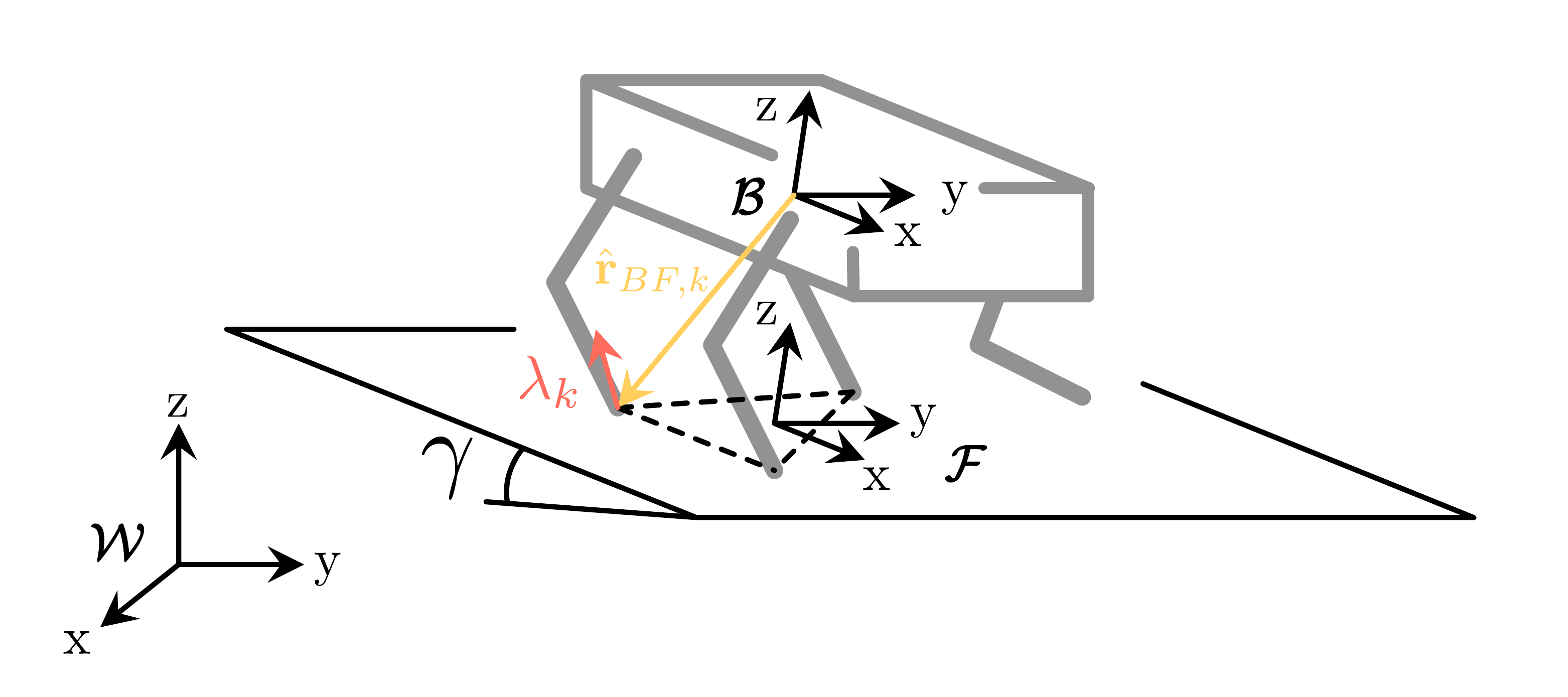}
    \caption{Definition of world frame $\pazocal{W}$ , footprint frame $\pazocal{F}$, and base frame $\pazocal{B}$.}
    \label{fig:control_frames}
\end{figure}

For the robot to remain stable on steep inclinations, its desired pose and foot placement must be adapted depending on the slope. Thus, an estimate of the underlying terrain orientation is needed. We use a similar approach as described in \cite{gehring2015dynamic}. The terrain is modeled as a plane whose parameters are estimated based on proprioceptive measurements. For convenience, additional coordinate frames are introduced as seen in Figure \ref{fig:control_frames}. The world frame $\pazocal{W}$ is an inertial frame; the base frame $\pazocal{B}$ is fixed to the torso of the robot and lies in its center of gravity. Its x-axis is aligned with the heading direction of the torso, while the y-axis points in the lateral direction. The footprint frame $\pazocal{F}$ lies in the mean position of all feet that are in ground contact. The x-y-plane of this coordinate system is parallel to the estimated terrain plane, with x pointing in the heading direction.

We aim to find the orientation of the terrain with respect to the world frame. This can be expressed as the rotation from $\pazocal{F}$ to $\pazocal{W}$, which is given by 

\begin{equation}
\label{eq:orientation_dependency}
    C_{\pazocal{W}\pazocal{F}} = C_{\pazocal{W}\pazocal{B}}C_{\pazocal{B}\pazocal{F}} \text{ .}
\end{equation}

$C_{\pazocal{W}\pazocal{B}}$ can be directly extracted from the onboard IMU. The IMU data is solely used for locomotion control where drifting did not pose an issue. For navigation and respective localization tasks, IMU data would need to be fused with leg kinematics \cite{bloesch2013state} and visual data \cite{wisth2019robust} to avoid error accumulation in the state estimation. To estimate $C_{\pazocal{B}\pazocal{F}}$, we use the foot positions in the base frame, calculated from encoder data. A plane is fitted through all feet' contact points. The rotation from the footprint frame to the base frame can then be extracted as the rotation difference between the $z$-axis of the footprint frame and the plane's normal vector.
The estimate of the terrain orientation is used to position the torso in a way that the normal forces at the feet in ground contact are evenly distributed. 

This is accomplished by moving the torso position such that the projection of the base frame origin lies on the $y$-axis of the footprint frame. Due to the two DOF legs, the roll angle is not adapted to the terrain to increase stability. Therefore, the robot's pitch orientation is set to be parallel to the footprint frame while the roll orientation stays horizontal w.r.t the world frame.

\subsubsection{Walking control}
To realize a walking motion, \textit{SpaceBok} employs a locomotion control architecture composed of several planning and control elements.
First, a leg coordinator assigns swing and stance legs according to a predefined hand-tuned gait pattern (Sec.~\ref{sec:gaitselection}). The switching from stance to swing is done purely based on timing. This simple strategy can lead to hard impacts on firm ground if the legs switch from swing to stance too early or too late. However, on granular media, the impact is absorbed by the ground, and the strategy proved to be surprisingly robust. Furthermore, contact detection based on sensor data would be more demanding on such a terrain due to the uncertainty of the soil properties. 
During the swing phase, the foot trajectories are planned in the hip frame of the respective leg based on a predefined step height and a step length according to the desired heading velocity. The Cartesian-space positions of the trajectory are mapped to joint positions via inverse kinematics and tracked by a joint-space proportional-derivative (PD) controller of the form
\begin{equation}
\boldsymbol{\tau}^*=k_p(\mathbf{q}_j^*-\mathbf{q}_j) - k_d(\mathbf{\dot{q}}_j^*-\mathbf{\dot{q}}_j) \text{ ,}
\end{equation}
where $\boldsymbol{\tau}^*$ denotes the desired actuator torques, $k_p$ and $k_d$ the proportional and derivative gains, $\mathbf{q}_j$ and $\mathbf{\dot{q}}_j$ the joint angles and velocities and $\mathbf{q}_j^*$ and $\mathbf{\dot{q}}_j^*$ the respective desired values obtained from the foot trajectory planner.

The desired torso pose and twist during stance is controlled with a Virtual Model Controller (VMC) \cite{pratt2001vmc}. The virtual force $\boldsymbol{F_v} \in R^3$ controls the robots position and velocity in a PD fashion towards the desired positions $r_{x}^*$, $r_{y}^*$ and $r_{z}^*$ and velocity $v_{x}^*$, which are defined in the foothold frame. Furthermore, the gravitational force acting on the robot of mass $m$ is compensated, which results in the equation
\begin{align} 
\label{eq:virtual_force_calculation}
\boldsymbol{F_v} = \begin{bmatrix}
f_x\\ 
f_y\\
f_z
\end{bmatrix} =
\begin{bmatrix}
 k_{p,x}^f(r^*_{FB,x} - r_{FB,x}) + k_{d,x}^f(v^*_{B,x} - v_{B,x}) \\ 
  k_{p,y}^f(r^*_{FB,y} - r_{FB, y}) - k_{d,y}^f \cdot v_{B,y} \\ 
 k_{p,z}^f(r^*_{FB,z} - r_{FB,z}) - k_{d,z}^f\cdot v_{B,z}
\end{bmatrix}
+ C_{\pazocal{F}\pazocal{W}}
\begin{bmatrix}
 0 \\ 
0 \\ 
m \cdot g
\end{bmatrix} \text{ .}
\end{align}
Besides the virtual force, a virtual torque $\boldsymbol{T_v} \in R^3$ is calculated to correct the orientation error of the robot. The desired virtual torque controls the robot orientation expressed by the unit quaternion $\boldsymbol{p}_{WB}$ to the desired value $\boldsymbol{p}_{WB}^*$, which is obtained as described in Sec.~\ref{sec:terrainestimation}. Furthermore, damping is added by counteracting the angular velocity $\boldsymbol{\omega}_{WB}$. Thus, the calculation of the virtual torque reads

\begin{align} 
\label{eq:virtual_torque_calculation}
\boldsymbol{T_v} = k_p^T \cdot (\boldsymbol{p}_{FB}^* \boxminus \boldsymbol{p}_{FB}) - k_d^T  \cdot \boldsymbol{\omega}_{FB} \text{ .} 
\end{align}

The calculated virtual force is first transformed into the base frame and projected onto the x-z plane of the base frame, since the robot is unable to exert forces in y-direction of this frame. 

The virtual wrench is then mapped to foot forces using constrained quadratic optimization similar to \cite{gehring2016practice}. The formulation of the problem with $k$ feet in ground contact is:
\begin{align}
	\underset{\boldsymbol{x}}{\text{min}} \quad &||(\bm{Ax}-\bm{b})||_{2}^{2} \\
	\text{s. t.} \quad &F_{k}^n \geq F_{min}^n \label{eq:normal_forces_constraint}\\
	-&\mu F_{k}^n \leq F_{k}^t \leq \mu F_{k}^n \label{eq:tangential_forces_constraint} \\
	-&\boldsymbol{\tau}_{max} \leq \bm{J}^T\bm{x} \leq \boldsymbol{\tau}_{max} ,\label{eq:max_torque_constraint}
\end{align}
where
\begin{align}
\vspace{8pt}
\bm{A}&=\begin{bmatrix}
		\boldsymbol{I} & \boldsymbol{I} & ... & \boldsymbol{I} \\
		\boldsymbol{\hat{r}}_{BF,1} & \boldsymbol{\hat{r}}_{BF,2} & ... & \boldsymbol{\hat{r}}_{BF,k} 
    \end{bmatrix} \text{: Transformation matrix} \nonumber \\ 
\vspace{8pt}	    
\boldsymbol{x}&= \begin{bmatrix}
	    \boldsymbol{\lambda}_{1} \\
	    \boldsymbol{\lambda}_{2} \\
	    \vdots \\
	    \boldsymbol{\lambda}_{k}
	\end{bmatrix}\in R^{2k} \text{: 2D contact forces of the feet} \nonumber\\	
\vspace{8pt}	    
\bm{b}&=\begin{bmatrix}
	    \boldsymbol{F}_{v}  \\
	    \boldsymbol{T}_{v}
	\end{bmatrix}\in R^6 \text{: Virtual wrench acting at the COM} \nonumber  \text{ .}
\end{align}
In this formulation, $F_{k}^n$ and $F_{k}^t$ denote the normal (superscript n) and tangential (superscript t) contact force of leg $k$, $\boldsymbol{\tau}$ the stacked actuator torques, $\bm{J}$ the stacked actuator Jacobian, $\boldsymbol{I}$ the Identity matrix and $\boldsymbol{\hat{r}}_{BF,k}$ the position vector from robot base (COM) to each foot. The linear constraints of the optimization problem ensure positive normal components of the contact forces (Eq. \ref{eq:normal_forces_constraint}), the limitation of tangential forces to stay within friction cones with friction coefficient $\mu$ (Eq. \ref{eq:tangential_forces_constraint}), and the limitation of maximum available actuator torque ${\tau}_{max}$ (Eq. \ref{eq:max_torque_constraint}). The friction coefficient determines the upper limit for tangential forces to be applied to the ground. Thus, to give the constrained optimization the highest possible freedom and certainty that the motion will be executed correctly,  the feet have to be designed for high traction.

The yaw angle of the base is not taken into account in the optimization, which means that we do not apply a virtual torque in this direction. The reason is that due to the missing adduction/abduction DOF, yaw turning can only occur if the contact feet are slipping. However, this conflicts with the no-slipping constraint in Eq.~\ref{eq:tangential_forces_constraint}. In order to turn, we add an offset to the tangential foot contact force in the heading direction, where the components of the left and right legs are anti-parallel. This leads to slight slipping of the feet and enables the robot to turn despite the constraining leg topology. Since modeling slipping behavior is extremely hard, the linear mapping between the desired turning rate and offset force was hand-tuned. Accordingly, the contact forces are updated as
\begin{align} 
\boldsymbol{x}_{turn}&= \begin{bmatrix}
	    \boldsymbol{\lambda}_{1, turn} \\
	    \boldsymbol{\lambda}_{2, turn} \\
	    \vdots \\
	    \boldsymbol{\lambda}_{k, turn}
	\end{bmatrix}
	\qquad \text{with }
\boldsymbol{\lambda}_{i, turn} = \boldsymbol{\lambda}_{i} +-1^i \cdot k_{turn} \cdot \begin{bmatrix}
    \dot{\psi}^* \\
	    0
	\end{bmatrix}
\end{align}
where $\boldsymbol{x}_{turn}$ denotes the stacked contact forces with the added force in tangential direction, $\dot{\psi}^*$  the desired yaw angle rate, and $k_{turn}$ the turning gain. The iterator \textit{i} is even for right legs and odd for left legs.
The desired joint torques $\boldsymbol{\tau}^{*}_{VMC}$ during stance are finally acquired by using the Jacobian-transpose mapping of the desired foot forces 
\begin{align} 
\boldsymbol{\tau}^{*}_{VMC} = \bm{J}^T \boldsymbol{x}_{turn} \text{ .}
\end{align}

Soft-terrain adaptations to the contact force control, as, for example, proposed in \cite{lynch2020yieldterrain}, are not taken into account. Instead, we focus on adapting the foot hardware to match all sorts of compressible terrain. For a more in-depth description of the system's control architecture, please refer to earlier work \cite{arm2019icra}.

\section{Foot design for walking on planetary soils}\label{sec:spacebok_feet_design}

\subsection{Robotic feet fundamentals}
Like wheels of a rover, the design of specialized feet, which allow the robot to traverse loose, granular media, is of high importance for legged robots intended to walk on other planets. Nevertheless, leveraging large contact areas or grousers, which are extensively studied for wheeled planetary exploration rovers (e.g., \cite{moreland2012grousers, sutoh2012travelingperformance, inotsume2019wheeldesign}), have been barely touched in the research on legged robotics.

The majority of quadrupedal and multi-legged robots on Earth have passive point- or line contact feet since they provide sufficient points of contact with the environment.  Bipedal robots, in comparison, rely on actively controlled planar feet to exert torques on the ground for balance (i.e. \textit{Walk-Man} \cite{tsagarakis2017walkman}, \textit{Atlas} \cite{kuindersma2016optimization}, \textit{Cassie} \cite{gong2019cassie}, \textit{HRP3} \cite{kaneko2008hrp3} or \textit{Talos} \cite{stasse2017talos}). Interestingly, \textit{Cassie} was tested with human shoes during a sand-walking experiment, which qualitatively improved the performance\footnote{https://www.youtube.com/watch?v=VoD7hbssu-M}.

The disadvantage of point feet is their small surface area and the resulting high ground pressure. While not critical on hard surfaces, this becomes problematic on compressible, granular soil. Here, large ground pressure generally causes high sinkage, which may lead to stumbling and falls of the robot, as we have encountered in previous work \cite{kolvenbach2018isairas}, and potentially increases the energy consumption since mechanical work on the soil is energetically expensive \cite{Lejeune2071}. The lateral shear of soil particles creates additional issues, as the feet can get trapped in sand. Another disadvantage of the point foot design is the moving and non-constant point of contact within the soil, making contact detection and force control very challenging. Very few specialized, passive-adaptive planar feet have been developed in the past. Our group pioneered a sensor-equipped passive adaptive foot for \textit{ANYmal} \cite{kaeslin2018iros}. With this foot, we were able to double the surface area and thus reduce sinkage on granular media. The design incorporates Force/Torque sensors and IMUs to inspect and classify the soil in front of the robot \cite{kolvenbach2019ral}. While the foot showed a good performance in acquiring knowledge of the surroundings and less sinkage, it was not specifically optimized for sand- and slope-walking nor validated on representative soil. 
From previous studies, we concluded two main objectives that have to be achieved by the feet:
\begin{itemize}
\item The leg's sinkage into the soil should be minimized on all possible soils to avoid situations in which the foot gets stuck, or the leg reaches kinematic limits. Small footprints also increase the variance in penetration depths, which results in an indifferent contact state. Because of unclear penetration depths, the uncertainty during motion planning grows, and the risk for falls increases.

\item The applicable traction force shall be maximized to increase the robot's controllability and avoid energy loss due to slip. In the case of low traction and slippage, the robot cannot walk and might fall due to an imbalance of motion.
\end{itemize}

Additionally, we found that the design should remain fail-proof and simple. Since feet experience repeating impacts and interactions with the environment, the use of complicated mechanics or essential sensors for locomotion should be avoided to increase robustness. The feet should also remain lightweight since increased inertia of the leg due to heavy feet increases energy consumption and decreases motion tracking performance.

\subsection{Foot development and test setup}\label{sec:single_foot_test}

Evaluating the performance of foot designs on soil is a daunting task due to the complexity of the physical effects at play. In the past, several semi-empirical wheel-soil interaction models have been developed, such as the famous Bekker model and its derivatives \cite{bekker1956, bekker1960, rodriguez2019highspeed}. While those methods have been applied to planetary exploration systems with certain success, the modeling of leg-foot-soil interaction, in comparison, is more diverse and dynamic and has only become a subject of study recently \cite{ding2013}. Overall, while first-principles models exist, modeling the interaction between legged robots with soil remains one of the grand challenges in robotics \cite{yang2018challenges}.

In addition to the complexity of modeling the interaction, acquiring the required soil parameters from planetary bodies in the first place presents another challenge. Previous studies have proposed different terramechanical models that can be used to estimate the static sinkage of a wheel or foot into the lunar regolith. For example, \cite{carrier} developed a linear Winkler model that has been derived using measurements of Apollo astronaut bootprints. While this model is expected to provide realistic sinkage estimates for the known, flat, equatorial lunar surface, it cannot be applied to slopes and the yet untraversed regions of the Moon, such as pyroclastic deposits and the polar permanently shadowed regions, both high priority targets for future exploration missions. In order to address this lack of knowledge, \cite{bickel2019, bickelandkring}, and \cite{sargeant2020} developed a purely remote sensing-based method that uses boulder tracks and Hansen's bearing capacity theory \cite{hansen} to derive first-order estimates of the surface and sub-surface strength of these unknown regions. While this method provides insight into untraversed surfaces' general bearing behavior, its results are not expected to be exceedingly accurate due to the uncertainties associated with the method.

Because of the complexity of the interaction and relative uncertainty of modeling the feet-soil behavior, we choose an empirical approach to determine best practices for the design of \textit{SpaceBok}'s feet. While this approach only provides trends and probabilistic outcomes, it allows us to gain insight into the performance drivers for our feet. A challenge in empirical testing is to find a trade-off between A) the broadness of the scenario (which has to be general enough to answer a hypothesis) and B) the number of tests required. Additionally, also empirical tests may fall short in capturing all phenomena encountered in the real world, such as, in our specific case, the effect of reduced gravity on the soil and the behavior of the robot. Until now, the correlation between the gravitation and repose angle of granular materials is still unclear \cite{niksirat2020slip, chen2019gravity}.

We found that to derive viable design guidelines, we require 1) testing on various regolith simulants with a broad range of geomechanical properties and 2) testing on flat and inclined surfaces. We find that a certain level of flotation "over-performance" is desired to enable \textit{SpaceBok} to deal with local surface strength anomalies, i.e., unexpected, locally constrained, and particularly poor soil conditions.

\begin{figure}[!tb]
	\centering
	\begin{minipage}{.51\textwidth}
        \centering
 \includegraphics[scale=0.29]{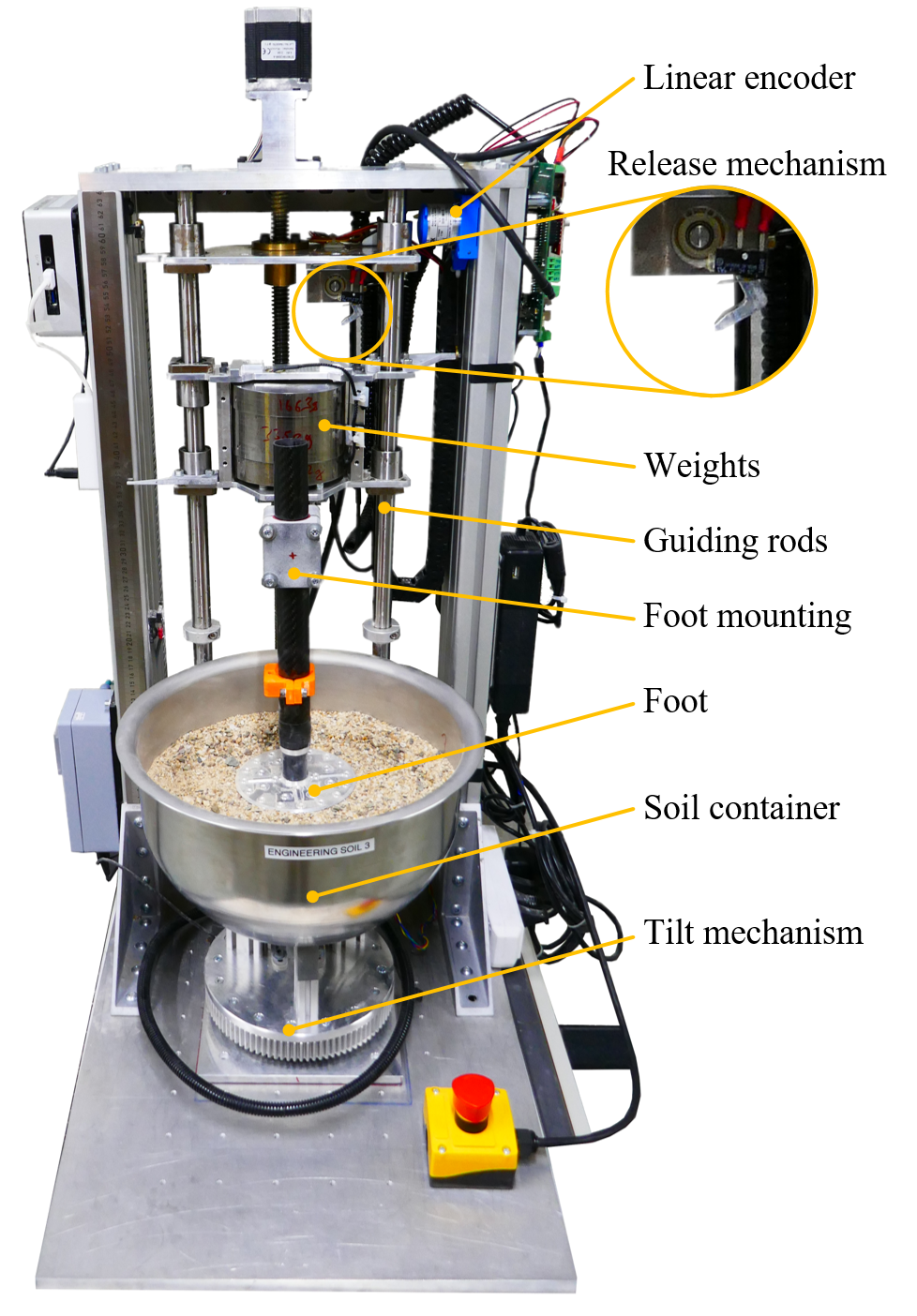}
    \subcaption{Depiction of the single-foot test setup}
    \label{Fig:singlefoottestbed}
    \end{minipage}%
    \begin{minipage}{.42\textwidth}
        \centering
        \includegraphics[width=0.78\textwidth]{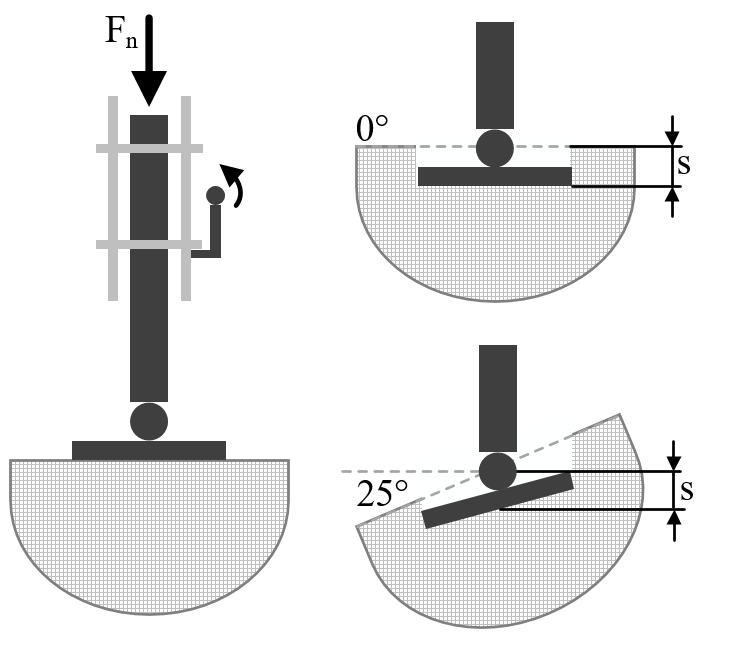}%
	    \subcaption{Sinkage test}
	    \label{Fig:sinkagetestsetup}
   	    \includegraphics[width=0.46\textwidth]{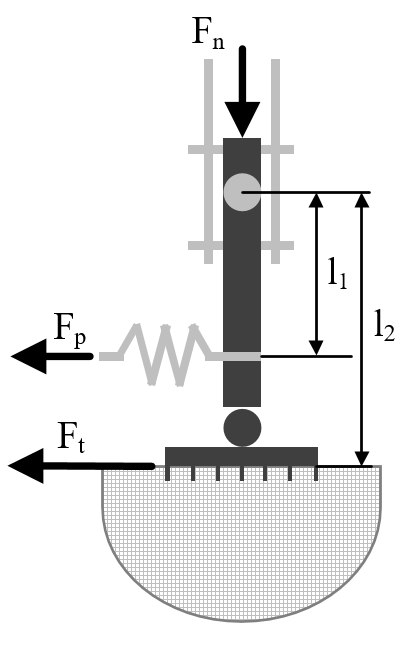}%
	    \subcaption{Traction test}
	    \label{Fig:tractiontestsetup}
    \end{minipage}
	\caption{Single-foot experiments are performed on a dedicated testbed a). The two test cases, namely a sinkage test on 25\degree~inclined and non-inclined soil are depicted in b), while the traction test case is depicted in c). }
	\vspace{-8pt}   
\end{figure}

We adopted a two-stage testing approach to determine the influence of the design parameters on sinkage and traction, and provide a relative comparison between our prototypes. The first test, referred to as the \textit{Sinkage test}, aims to determine the soil penetration depth (sinkage or flotation) of adaptive plates with a varying diameter on flat and inclined terrain. The second test, referred to as the \textit{Traction test}, aimed to determine the influence of material and grouser design on the traction performance, i.e., the foot resistance against slippage. To perform testing in a repeatable manner, we utilized a customized single-foot testing machine (Figure \ref{Fig:singlefoottestbed}). The testbed is equipped with actuators to move the mounted experimental foot along the vertical axis. A release mechanism allows for the precise dropping of the foot while an attached linear encoder reports the sinkage into the soil. The exchangeable soil container is mounted on a tilt mechanism that allows for the recreation of slopes up to 25\degree. We used representative lunar and martian soil simulants during our experiments, acquired from RUAG Space and the Technical University of Braunschweig. \\

To guarantee uniform soil conditions throughout all tests, all samples were standardized by 1) soil raking and 2) placing them on a shaking table for 60 seconds (model 'Spartan', vibratory sieve shaker analysette 3, Fritsch GmbH). This specific duration of shaking has been chosen as a best-effort approach to recreate the expected in-situ bulk densities of these soils on the martian and lunar surface, as determined by past robotic and human exploration missions \cite{linke2018tubs, brunskill2011}. Table \ref{Tab:soils} summarizes the specifications of the soils. We repeated each test case at least three times, which resulted in 142 sinkage and 101 traction tests.

\renewcommand{\multirowsetup}{\centering} 
\setlength{\tabcolsep}{2pt}

\begin{table}[!tb]
\caption{The properties of the analog soils used in this study.}
\label{Tab:soils}
\begin{center}
\begin{threeparttable}
\begin{tabular}{p{3.2cm}|p{3.2cm}|p{3.2cm}|p{3.2cm}|p{3.2cm}}
\toprule
  & \textbf{ES-1} 
 & \textbf{ES-2}
 & \textbf{ES-3}
  & \textbf{TUBS}
 \\
\midrule
Appearance & Very ﬁne-grained, well-sorted, poorly-graded, porous, highly compressible, and collapsible soil. & Silty to very ﬁne, well-sorted, poorly-graded sand. & coarse, poorly-sorted, well-graded sand with a prominent silty component. &  A poorly-graded, well-sorted, silty, highly-cohesive, very fine sand. \bigstrut \\

Classification\tnote{*} & “silty clay” (ML) or a “lean clay” (CL) & “silty sand” or “sandy silt” (ML) & “silty sand” (SM) & “silty sand” (SM) \bigstrut \\

Occurrence\tnote{**} & A thin aeolian dust deposit that occasionally occurs on top of other surface materials on Mars.  &  A common aeolian, sandy deposit on the surface of Mars, forming geomorphic features such as dunes and sand ripples. & martian scree, coarse aeolian sands, and polymodal surﬁcial lag. Coarse scree and aeolian accumulations occur in terrain with escarpments. &  Regular lunar mare or highland terranes. Pyroclastic and shadowed regions likely have different soil properties \cite{bickel2019, sargeant2020}. \bigstrut \\

Modal grain size \tnote{**} & $ \sim \SI{10}{\um}$ & -  & $ \sim \SI{400}-\SI{600}{\um}$ & $ \sim \SI{60}-\SI{80}{\um}$  \\

Min. grain size \tnote{**} &$ < \sim \SI{10}{\um}$ &  $> \sim \SI{30}{\um}$  &   $> \sim \SI{30}{\um}$  & $\sim \SI{10}{\um}$ \\

Max. grain size\tnote{**} &  $\sim \SI{32}{\um}$ &  $\sim \SI{125}{\um}$ &  $\sim \SI{20000}{\um}$ & $\sim \SI{200}{\um}$ \\

Int. friction angle\tnote{*} & $34^{\circ}\pm4^{\circ}$ & $37^{\circ}\pm5^{\circ}$  & $35^{\circ}\pm5^{\circ}$  & $43.85^{\circ}\pm1.95^{\circ}$ \\

Cohesion\tnote{**} & $1\pm\SI{0.5}{kPa}$  &   $0.75\pm\SI{0.75}{kPa}$ &   $0.15\pm\SI{0.15}{kPa}$  & $0.51\pm\SI{0.8}{kPa}$  \\

Nom. Bulk density\tnote{**}  &    $1.2\pm\SI{0.2}{g/cm^{3}}$ &  $1.5\pm\SI{0.1}{g/cm^{3}}$  &   $1.6\pm\SI{0.1}{g/cm^{3}}$  &  $\SI{1.527}{g/cm^{3}}$ \\

Exp. Bulk density\tnote{***}  &  \SI{0.88}{g/cm^{3}} & \SI{1.59}{g/cm^{3}}  &  \SI{1.68}{g/cm^{3}}  & \SI{1.56}{g/cm^{3}} \\

\bottomrule
\end{tabular}
\begin{tablenotes}
\item[*] According to USCS ASTM D2487 standard 
\item[**] Taken from \cite{michaud2010isairas}, \cite{winnedael2014technicalnote}, \cite{linke2018tubs} 
\item[***] Experimentally verified
\end{tablenotes}
\end{threeparttable}

\end{center}
\end{table}

\begin{figure}[!tb]
	\centering
	\begin{subfigure}[t]{0.38 \textwidth}
   		\includegraphics[width=\columnwidth]{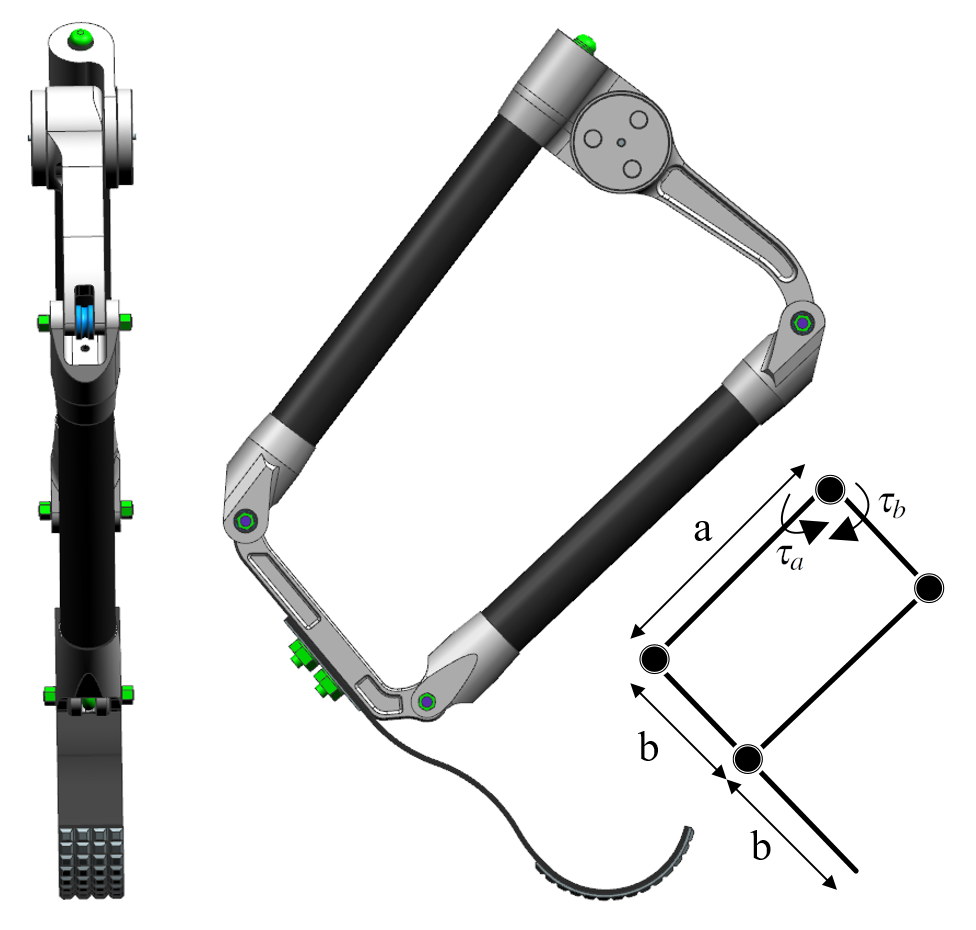}
   		\caption{The original carbon foot.}
   		\label{fig:carbon_kinematics}
	\end{subfigure}
	\hfill
	\begin{subfigure}[t]{0.57 \textwidth}
   		\includegraphics[width=\columnwidth]{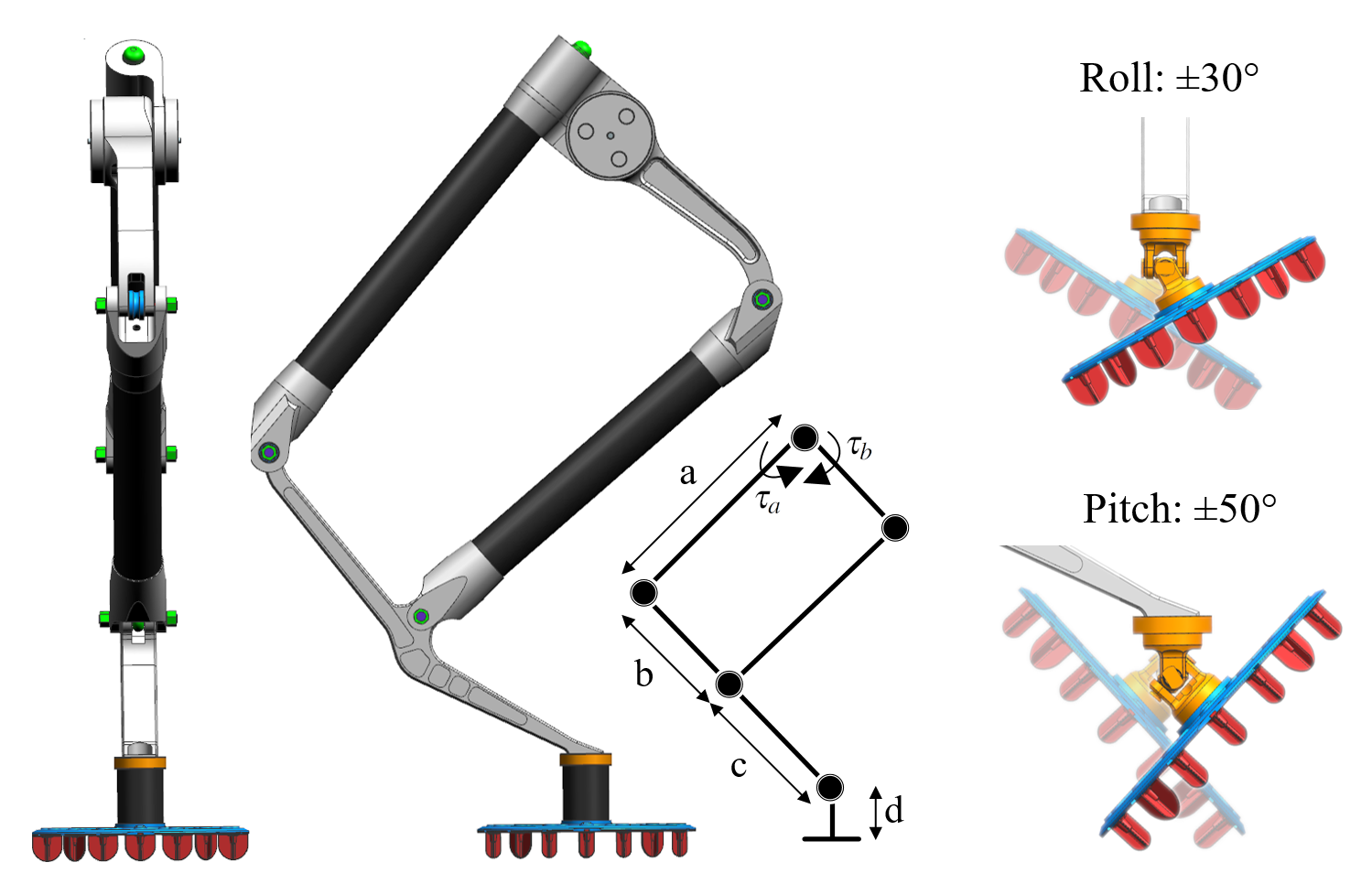}
        \caption{The passive-adaptive planar foot.}
        \label{fig:planar_kinematics}
	\end{subfigure}
	\caption{Schematic view of the leg kinematics on \textit{SpaceBok}. a) shows the original design while b) shows the updated foot design with adaptive ankle joint. The linkage lengths are as follows a=\SI{250}{mm}, b=\SI{120}{mm}, c=\SI{130}{mm} and d=\SI{20}{mm} (center of universal joint to bottom of planar surface).}
	\label{fig:leg_kinematics}
      \vspace{15pt}     
\end{figure}

For the \textit{sinkage test}, we investigated four different surface areas: The original C-shaped foot of the robot with \SI{12}{cm^2} surface area in contact, and a small (\SI{30}{cm^2}), medium (\SI{70}{cm^2}) and big (\SI{110}{cm^2}) circular plate. The rigid C-shape foot is made of carbon and includes a rubber sole; the plates are 3D printed from Nylon (PA12). To enable a plane contact, we introduced two passive DOFs at the ankle so that the foot can adapt to the terrain. For this, we utilized a previously developed universal joint \cite{kolvenbach2020jfr} that allows the foot to passively adapt to the ground in the roll- and pitch orientation with a range of motion of \SI{\pm 30}{\degree} and \SI{\pm 50}{\degree}, respectively. A rubber tube surrounding the universal joint retracts the sole back to an initial configuration. We adapted the kinematic chain of the leg accordingly (Figure \ref{fig:leg_kinematics}). During the test, the feet were released from a rest position right above a soil container. To mimic a realistic contact interaction, the leg was pre-loaded with a weight of $F_n$=\SI{80}{N}, representing the average contact force during a static walk on earth. We tested the static sinkage of all foot designs and soils on flat terrain and 25\degree~inclined terrain (Figure \ref{Fig:sinkagetestsetup}), representing the "maximum traversable slope angle" for future surface exploration platforms as requested by the Lunar Exploration Science Working Group \cite{lex}. 

\begin{figure}[!tb]
\centering
\begin{minipage}{0.95\textwidth}
\centering
\subfloat[The planar feet used during the traction tests consist of a hollow 5mm stud (A), hollow 10mm stud (B), hollow 15mm stud (C), rigid 10mm stud (D) and the final design without (E) and with grousers (F).]{
\label{fig:testfeet}
\includegraphics[scale=0.9]{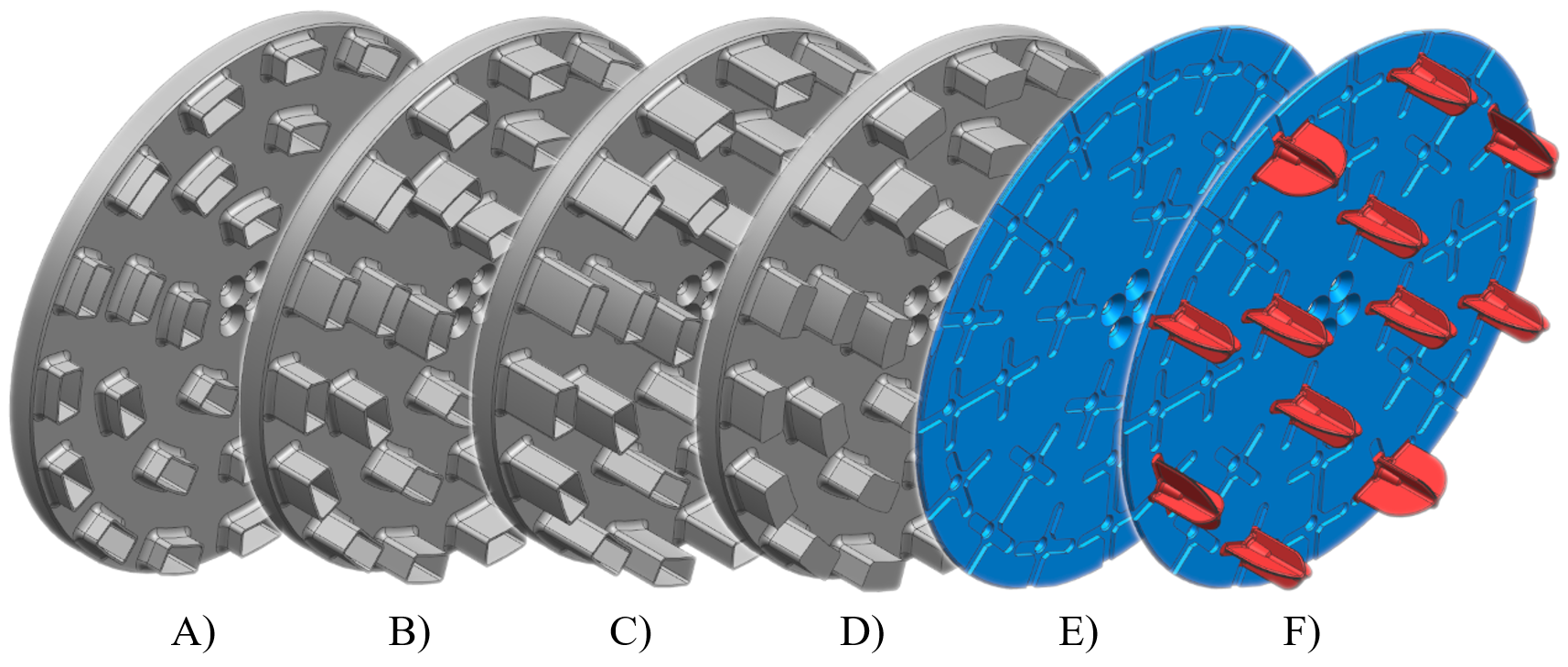}
}
\end{minipage}%
\hspace{1cm}
\begin{minipage}{\textwidth}
\centering
\subfloat[Detailed view of the stud design.]{\label{fig:studfoot}\includegraphics[scale=.95]{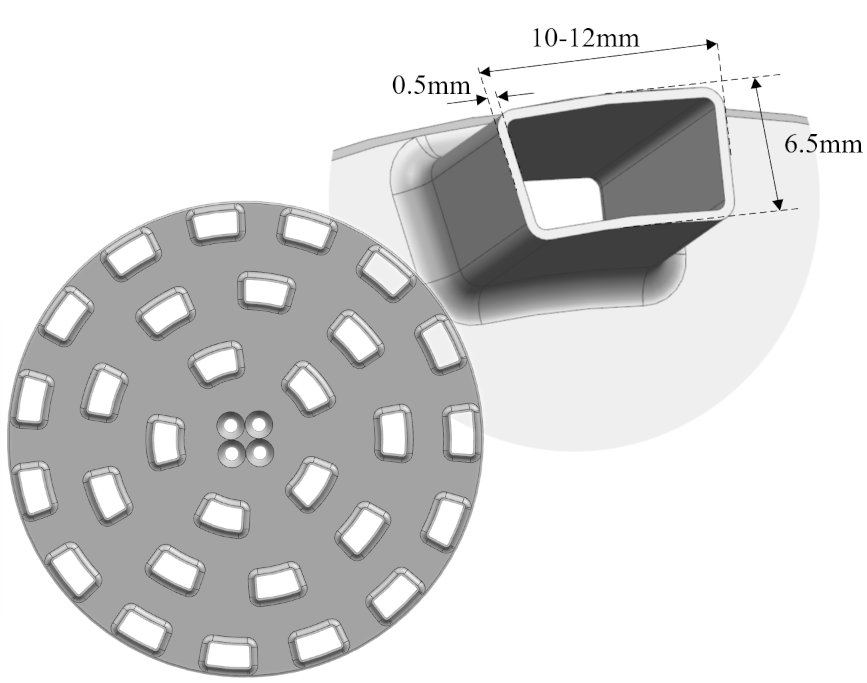}}
\centering
\subfloat[Detailed view of the final design.]{\label{fig:modularfoot}\includegraphics[scale=.95]{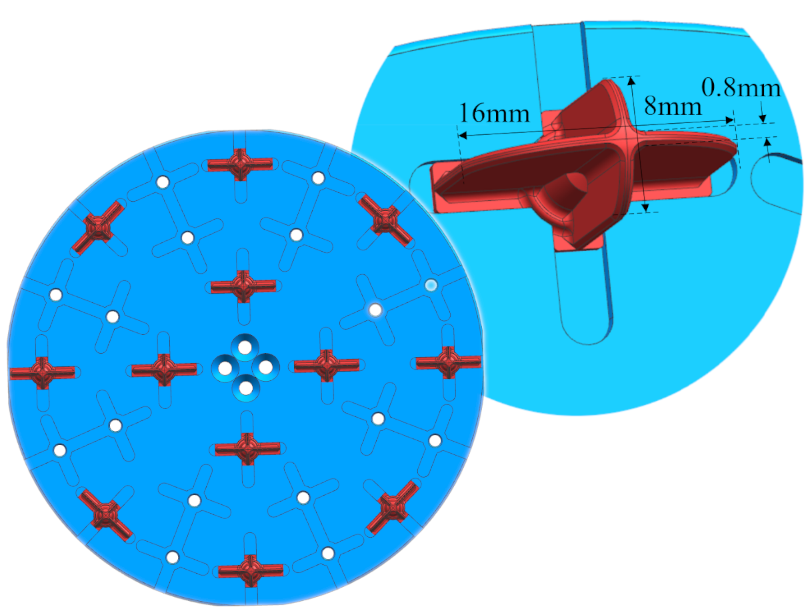}}
\end{minipage}\par\medskip
\caption{Detailed view of the various feet tested for their traction performance.}
\label{fig:feetdesigns}
\end{figure}

For the \textit{traction test}, we developed seven feet and five different grouser designs: The original rigid carbon foot, small hollow studs (\SI{5}{mm}), medium hollow studs (\SI{10}{mm}), large hollow studs (\SI{15}{mm}), medium rigid studs (\SI{10}{mm}), the final design without and with grousers (\SI{12}{mm}) attached (Figure \ref{fig:testfeet}). \\

Due to a lack of available research on the topic, we chose the grousers' lengths and spacing based on qualitative observations made by a study for wheeled exploration rovers \cite{inotsume2019wheeldesign}. The rectangular, hollow stud design was selected to provide sufficient structural integrity. Chevron grousers were not investigated since they proved worse performance on wheels on steep inclines compared to straight grousers \cite{inotsume2019wheeldesign}. The robot can walk forward and backward, so we designed the grouser symmetrically. 

We placed the pre-loaded foot on non-inclined soil during the test and applied a linearly increasing force via a spring parallel to the soil surface (Figure \ref{Fig:sinkagetestsetup}c). We used a load cell to record the maximum pulling force $F_p$ at which the foot started to slip. The traction coefficient $\mu_t$ was determined by taking the distance $l_1$ (from the foot attachment to spring attachment) and $l_2$ (from foot attachment to contact point) into account (Equation \ref{eq:traction_coeff}). We chose the foot's bottom plate as a contact point, similar to the performance evaluation of wheel-soil interaction \cite{rodriguez2019highspeed}. On the bedrock's hard surface, we used the distance from the foot mounting to the grousers' tip. Due to the technical limitation of the test setup, we could only test traction on horizontal surfaces. 

\begin{equation}
\label{eq:traction_coeff}
	\mu_t = \frac{max(F_t)}{F_n} = \frac{max(F_p) \cdot l_1 \cdot l^{-1}_{2}}{F_n}
\end{equation}

\subsection{Single-foot experimental results}

\subsubsection{Sinkage test result}

\begin{figure}[!tbh]
	\centering
	\begin{subfigure}[t]{0.48 \textwidth}
   		\includegraphics[width=\columnwidth]{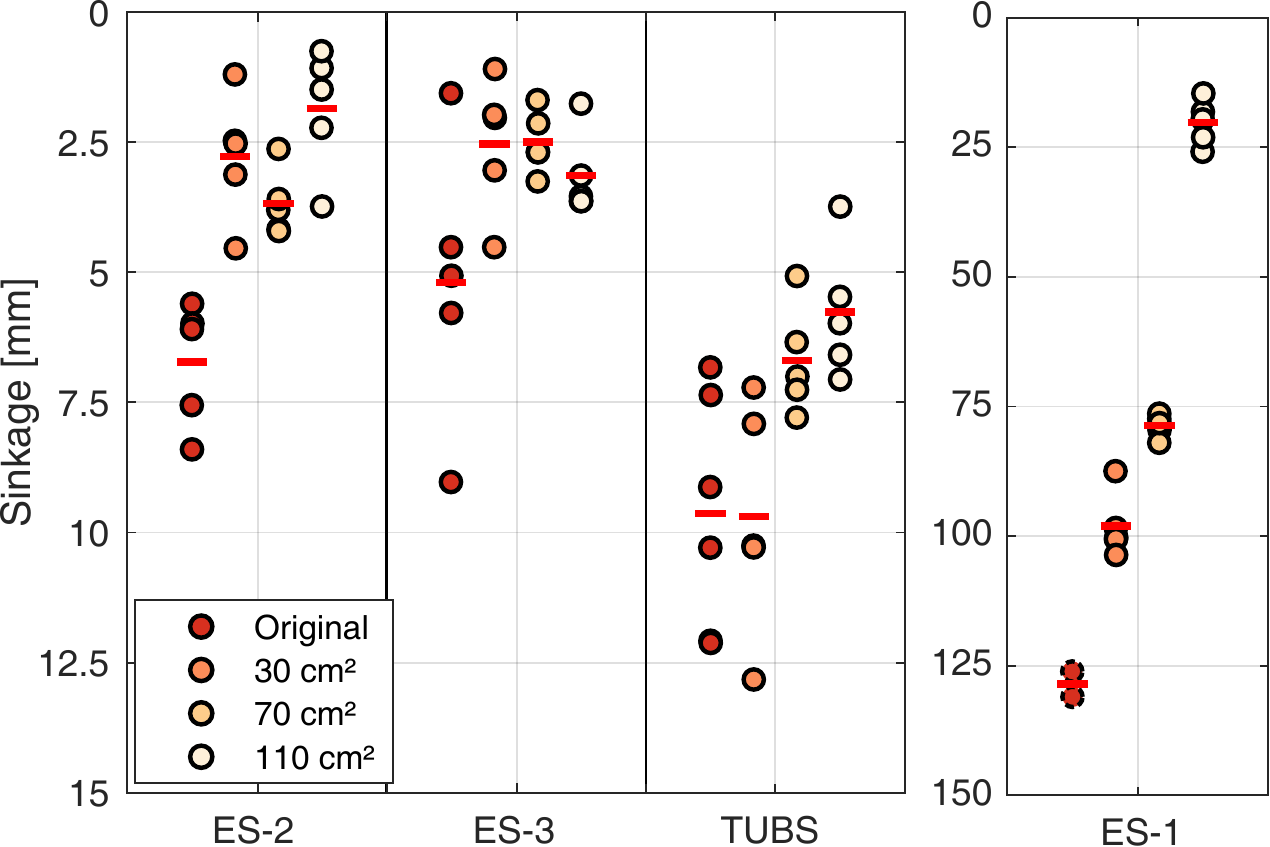}
   		\caption{Soil at \SI{0}{\degree}}
	\end{subfigure}
	\hfill
	\begin{subfigure}[t]{0.47 \textwidth}
   		\includegraphics[width=\columnwidth]{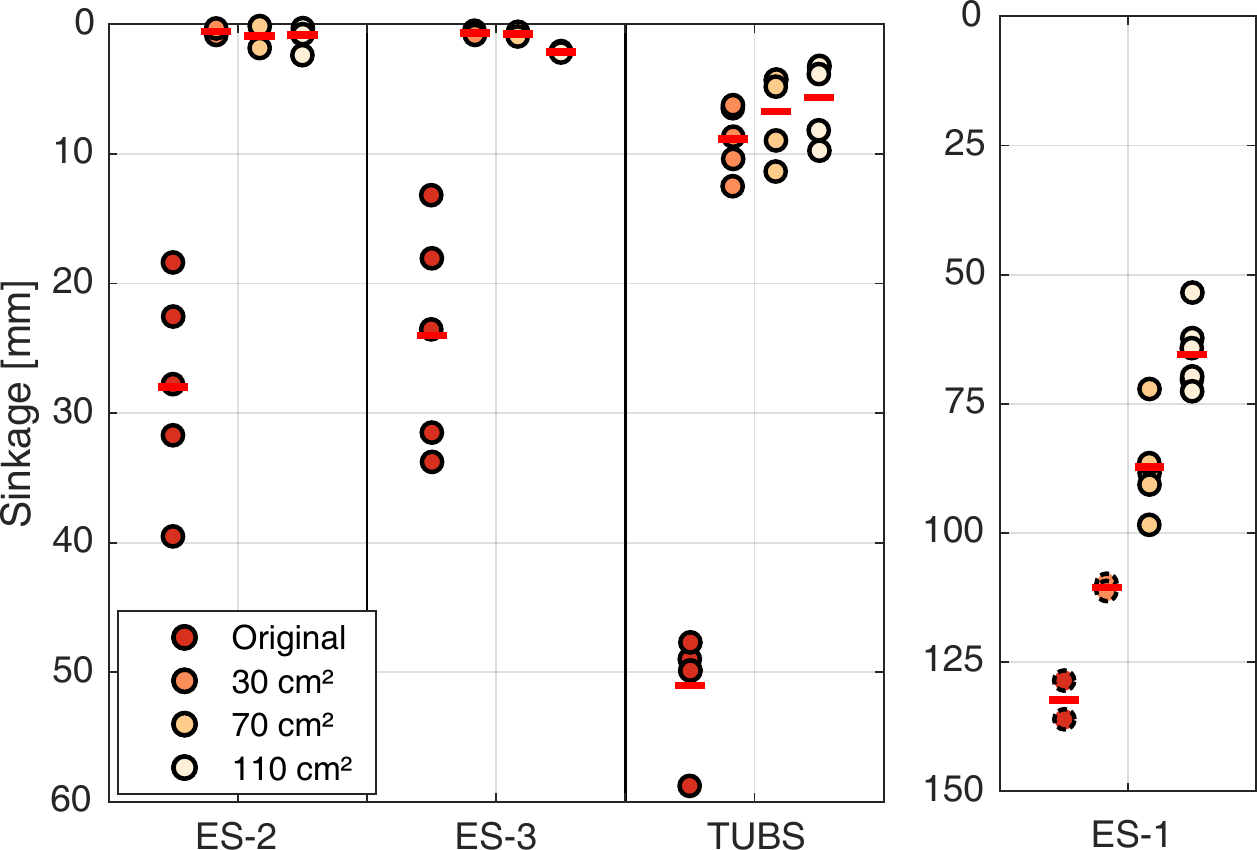}
        \caption{Soil inclined by \SI{25}{\degree}}
	\end{subfigure}
	\caption{Sinkage of the feet in various analog soils depending on surface area and soil inclination. The dashed circles on the \textit{ES-1} experiments indicate that the respective foot sank close to the bottom of the container where border effects influence the measurement. }
	\label{fig:sinkage}
        \vspace{-5pt}     
\end{figure}

The tests revealed insights into the foot-terrain interaction (Figure \ref{fig:sinkage}). The original C-shape carbon foot with the small surface area demonstrated the highest sinkage on almost all soils. Only on the lunar soil simulant, a similar sinkage is observed with the smallest diameter planar foot.

\textit{ES-1} is the most challenging soil, with an average sinkage being one order of magnitude above other simulants, followed by the lunar soil simulant, \textit{ES-2}, and \textit{ES-3} respectively. On flat terrain, \textit{ES-2} and \textit{ES-3} have high resistance against (punch, local, and general) shear, which limits the sinkage of the feet. Generally speaking, even a small diameter of \SI{30}{cm^2} provides enough surface area to limit the sinkage to \SI{10}{mm} on most simulants, while larger diameters only provide marginal improvement. Nevertheless, to avoid excessive sinkage or stalling (catastrophic sinkage up to the apogee of the nominal foot trajectory) on \textit{ES-1} type soils, a large diameter is necessary. 

\begin{figure}[!tbh]
	\centering
	\begin{subfigure}[t]{0.45 \textwidth}
   		\includegraphics[width=\columnwidth]{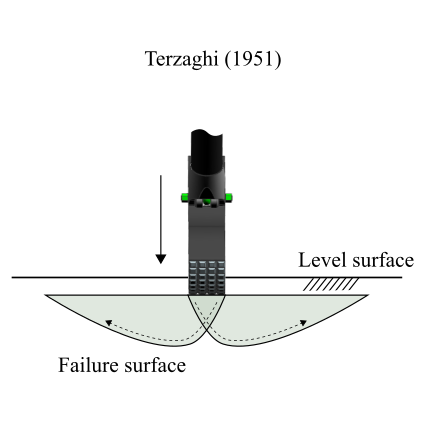}
   		\caption{Carbon foot on a level surface.}
   		\label{fig:terzaghi}
	\end{subfigure}
	\hfill
	\begin{subfigure}[t]{0.45 \textwidth}
   		\includegraphics[width=\columnwidth]{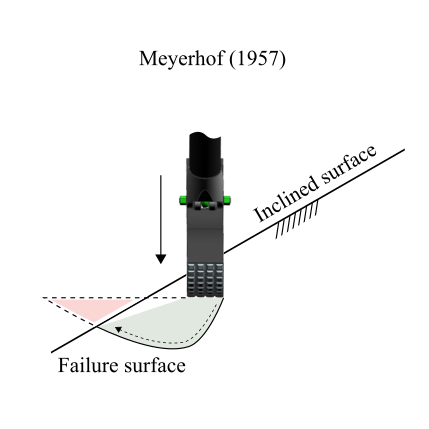}
        \caption{Carbon foot on an inclined surface.}
        \label{fig:meyerhof}
	\end{subfigure}
	\caption{The estimated failure surface for the carbon foot a) on level surface \cite{terzaghi} and b) on an inclined surface \cite{meyerhof1957ultimate}. With an assumed general shear mode (cross-section of foot and terrain). On inclined surfaces the reduced bearing soil volume leads to a reduced bearing capacity. Modified from \cite{bickelandkring}.}.
	\label{fig:shear_failure}
        \vspace{-5pt}     
\end{figure}

If the soil container is tilted by 25\degree, the overall bearing capacity of all tested simulants is significantly reduced, highlighting the importance of adaptation and larger foot diameters. Bearing capacity on slopes is generally reduced due to the reduced soil volume counteracting the bearing pressure \cite{meyerhof1957ultimate} (Figure \ref{fig:shear_failure}). The penetration of the C-shape foot is significantly higher on inclined slopes compared to the adaptive feet. Generally, the tests on inclined soil underline the importance of large foot diameters that can adapt to the slope to prevent excessive sinkage, particularly in less favorable soil.

\subsubsection{Traction test result}

\begin{figure}[!tb]
\centering
   \includegraphics[scale=0.68]{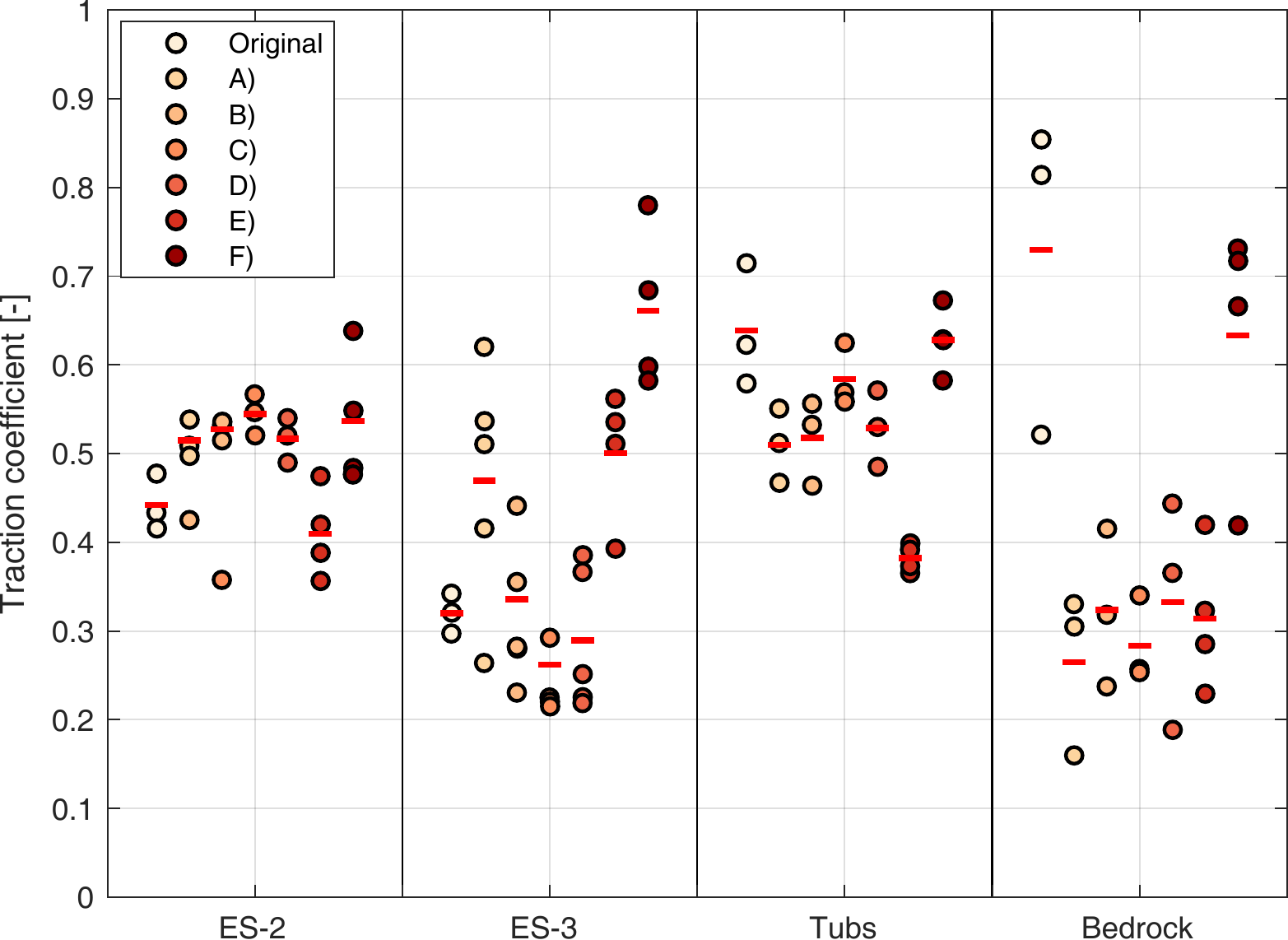}
    \caption{Traction coefficient on soil analogs for varying grouser/foot pad designs}
    \label{Fig:tractiontest}
\end{figure}

The traction tests on flat terrain revealed performance differences of the different grouser designs (Figure \ref{Fig:tractiontest}). On \textit{ES-2}, all stud designs perform equally well. In turn, the C-shape foot has trouble providing traction due to the soil's relatively high bearing capacity: as the sinkage of the C-foot is limited, the foot only activates the uppermost layer of the granular material, effectively reducing the traction. The heterogeneous, well-graded \textit{ES-3} is an interesting material since the small, \SI{5}{mm} studs perform relatively well, as the small studs can achieve full penetration, i.e., the foot sinks up to the bottom plate. The simulant's compression by the entire foot then appears to increase the traction as the soil cannot easily dilate upwards. In contrast, the larger, hollow, and rigid studs of 10-\SI{15}{mm} cannot penetrate the soil entirely despite the relatively thin stud design; this allows the soil to dilate upwards and effectively reduces the foot's traction performance on ES-3. Notably, the plain footplate without any studs (Type E) outperforms the foot with long studs. These observations highlight the fact that traction on granular media is mainly controlled by the sinkage of the foot and the overall shear resistance of the granular material.

The optimal foot would sink in just enough to provide maximum traction, i.e., would optimize the relation of sinkage and traction. We also observed blockages due to particle interlocking within the hollow studs on \textit{ES-3}. On the fine-grained \textit{TUBS} material, long studs provide an advantage over shorter studs as they activate a larger soil volume and thus provide higher resistance against horizontal pulling. The original carbon foot also performs well since it sinks deep in the soil, and thus, a larger force is required to shear the soil.

On bare bedrock, the C-shape foot is superior to the adaptive foot because of the large normal force and the high-friction rubber sole. The nylon-stone pairing and large surface area of the adaptive feet have a low friction coefficient in comparison. The foot with rectangular studs does not provide any advantage since its vertical walls do not engage with the rock. The surface area of the rectangular studs is level which makes it additionally difficult to interlock with the stone. We note that traction is rather controlled by macroscopic rock edges causing a dilation of the foot than the frictional properties of the rock mass. As chosen for the final design, a pointy design of the studs increases the interaction points with the rock and thus the traction. Pointy studs reach similar traction levels on average compared to the point foot design.

The planar feet \textit{E)} and \textit{F)} of Figure \ref{Fig:tractiontest} differ solely by the usage of grousers, and the data clearly highlights their benefits in terms of tractive performance. The most significant improvement is visible on bedrock, where grousers increase the traction coefficient by 125\% due to macroscopic interlocking. The beneficial effect of grousers is also visible on compressible soils, where the traction coefficient was raised by 27\% on ES-2, 22\% on ES-3 and ES-2 66\% on TUBS.

\subsection{Lessons learned and final design}

Based on the lessons learned from the experiments, we derived a final design with the following features:
\begin{itemize}
\item We chose a passive-adaptive planar foot design because it significantly reduced sinkage on inclined regolith simulants.
\item The largest surface area of \SI{110}{cm^2} provided the highest robustness against excessive sinkage in inclined simulants and prevented catastrophic sinkage in particularly unfavorable simulants, such as \textit{ES-1}. On \textit{ES-1}, the foot was able to guarantee foot floatation above the maximum of half the step height (\SI{75}{mm}). This choice represents our effort to select a design that is as robust as possible versus worst-case soil conditions, which caused mission fatalities in the past. In case soils with particularly challenging properties would be neglected (or could be excluded), foot designs with significantly smaller surface areas could be selected as well.
\item We selected rather long grousers of \SI{12}{mm} to improve the traction of \textit{SpaceBok} on soft soil. We limited the length of the studs to maintain a good balance between traction on soft (longer stud, more traction) and stiff soil (short stud, more traction). Similar results were found by \cite{inotsume2019wheeldesign} for wheels. 
\item The hollow stud design was converted to a closed star configuration since we found that large grains blocked the hollow studs on the heterogeneous ES-3 soil. Additionally, the star configuration reduces the surface area of the grouser tips, which eases soil penetration while providing a similar vertical surface, helping to optimize the trade-off between sinkage and traction. 
\item Pointy grousers allowed for macroscopic interlocking between grouser and a rigid material like bedrock, and we adopted this design.
\end{itemize}
The final design is weight-optimized (foot surface is made of aluminum) and robust (grousers are made of stainless steel). We selected the largest vertical surface area in the direction of travel.
In the following tests, we investigate the performance of the final adaptive-planar foot design and a rigid point contact foot on the robot.


\section{Slope walking on martian soil analog}\label{sec:ruag_tests}

\subsection{Deployment at RUAG Space}\label{sec:ruag_testsetup}

\begin{figure}[!b]
\vspace{-1em}
	\centering
	\begin{minipage}{0.7\textwidth}
        \centering
 \includegraphics[width=0.9\textwidth]{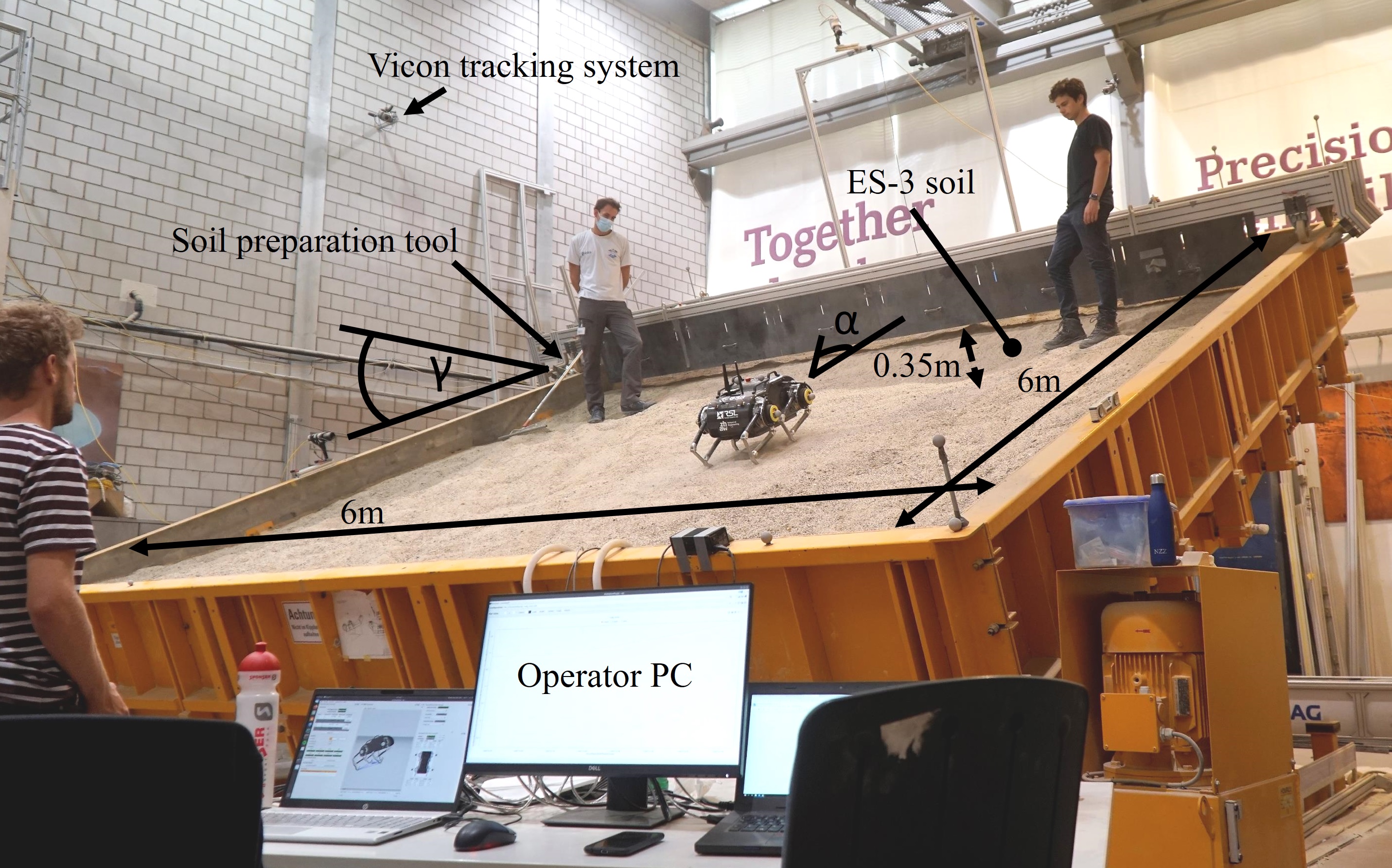}
    \subcaption{Depiction of the test setup at RUAG Space}
    \label{fig:ruag_testsetup}
    \end{minipage}%
    \begin{minipage}{.3\textwidth}
        \centering
   	    \includegraphics[width=\textwidth]{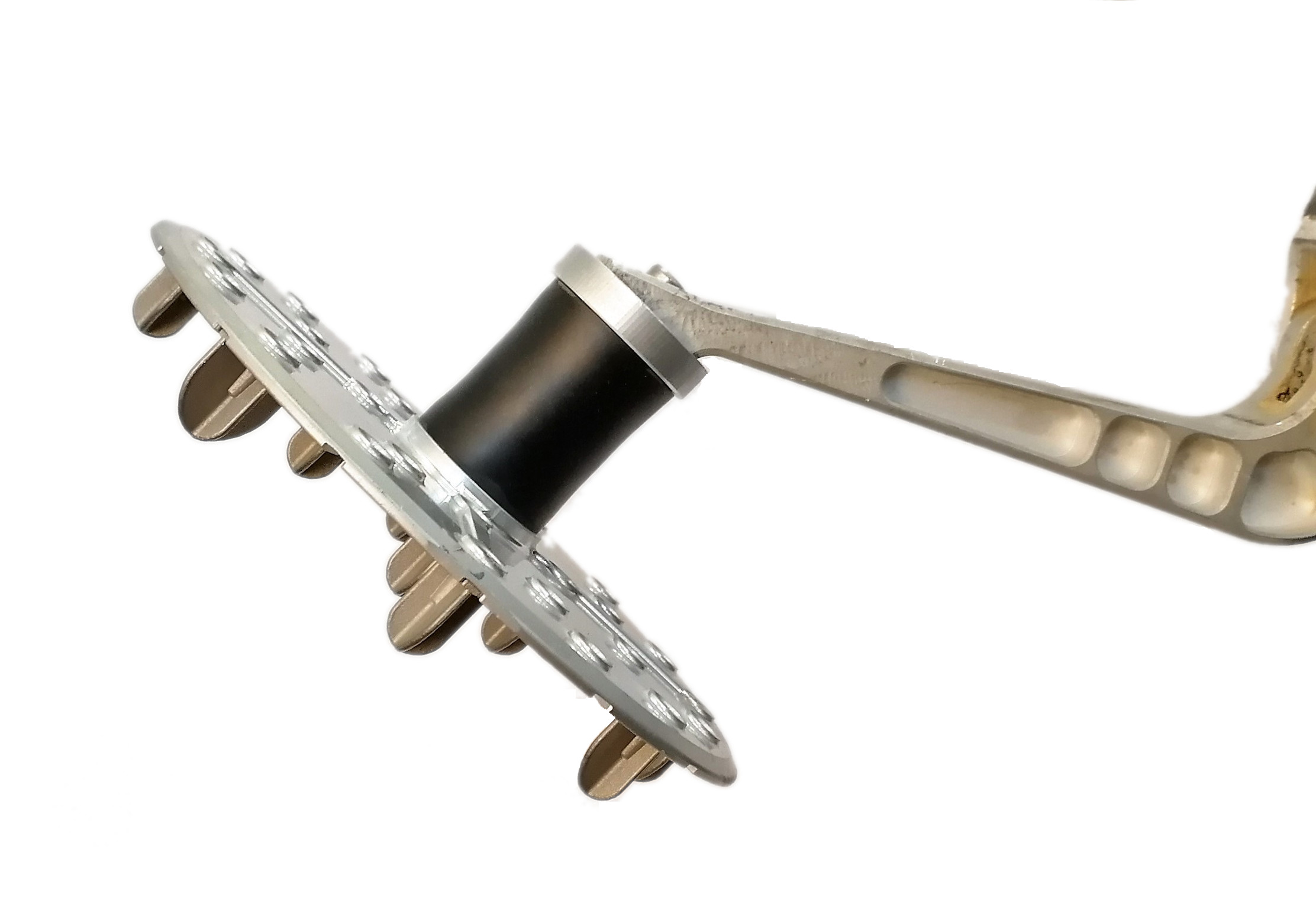}
	    \subcaption{Planar foot}
	    \label{fig:ruag_planarfoot}
	     \includegraphics[width=\textwidth]{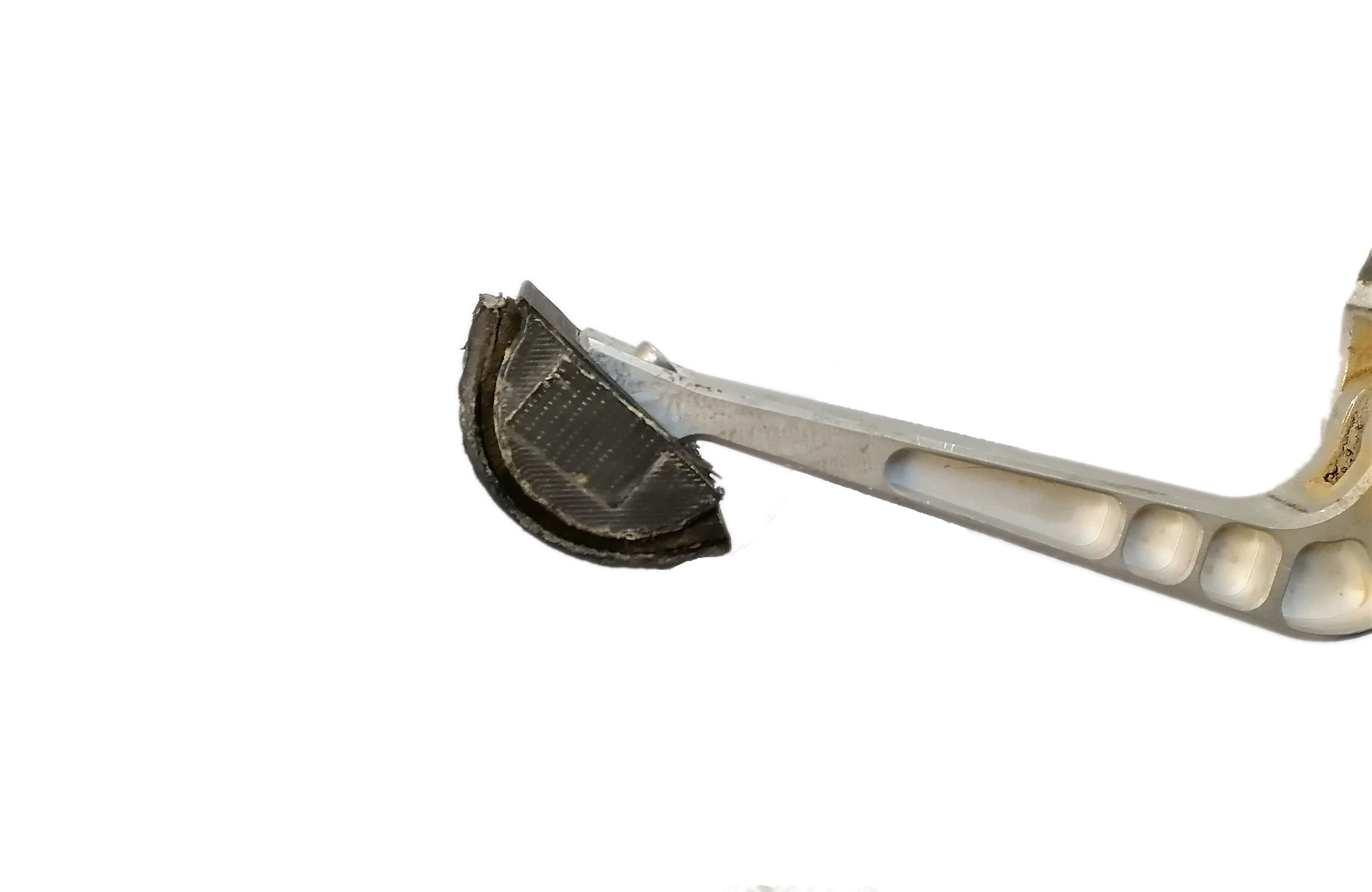}

	    \subcaption{Point foot}
	    \label{fig:ruag_pointfoot}
    \end{minipage}
	\caption{The experimental test setup during the field trial.}
\end{figure}

We tested the robot at the robotic test facility of Ruag Space, Zurich, which was designed for validation of the locomotion subsystem of ESA's ExoMars rover \textit{Rosalind Franklin}. The testbed consists of a \SI{6}{m} by \SI{6}{m} soil container filled with \SI{20}{t} of \textit{ES-3} (Figure \ref{fig:ruag_testsetup}). The testbed can be inclined from 0\degree~to 25\degree~with 0.1\degree~resolution. An absolute motion tracking system (Vicon) is installed in the facility and allows tracking of the robot with sub-centimeter accuracy. \\
We performed direct ascent and descent tests (Angle of attack (AOA) of 0\degree), during which we inclined the testbed in discrete steps to a slope of 0\degree, 5\degree, 10\degree, 15\degree, 20\degree, 22.5\degree, and 25\degree. The robot was not gravity-compensated during our experiments. We tested the final design of the planar foot (Figure \ref{fig:ruag_planarfoot}) as well as a point foot (Figure \ref{fig:ruag_pointfoot}). The original C-shape foot was replaced with a version that features the exact same foot shape and surface area but has a modified mount (identical to the planar foot mount) to allow for an easier and quicker exchange of the different foot types. We tested both feet with a static and a dynamic walking gait and repeated each test case at least three times to account for variations in soil preparation, yielding a total of more than 150 tests. Before each consecutive trial, the soil was loosened and leveled with a rake.  On average, the robot walked \SI{3}{m} in each test before reaching the end of the testbed. We also performed several diagonal ascents and descents on the 25\degree~inclined slope with an AOA of 30\degree, 45\degree, 60\degree, and 90\degree (parallel to the slope). Diagonal walks were repeated twice and lasted \SI{3.5}{m} on average. During all tests, the Vicon position data was acquired at \SI{80}{Hz}, the robot state at \SI{666}{Hz}, and the energy consumption via the BMS at \SI{3}{Hz}. 

\subsection{Slope walking experimental results}\label{sec:ruag_results}

\subsubsection{Walking performance on the slope}\label{sec:ruag_results_performance}
We validate the ability of the robot to ascend and descend slopes of up to 25\degree~(the maximum of the testbed) with the planar foot and the point foot using both the static walking gait and the trotting gait. Videos of the traverses are available online\footnote{https://youtu.be/VNPdlgvWWAY}. Up to a 20\degree~inclination, both feet can be used with both gaits without a significant decrease in performance.

As mentioned in Sec.~\ref{sec:gaitselection}, the gait parameters are hand-tuned for locomotion on granular media. A high foot apogee proves to work well in the experiments since it increases the chance of the feet being fully drawn out of the soil. This was especially important for the point feet, which sank deep into the sand. Furthermore, a lower CoG of the base increases the gait's robustness, although at the expense of higher energy consumption, as shown in Sec.~\ref{subsec:ruag_energy_consumption}. We use a foot apogee of \SI{0.15}{m} and \SI{0.1}{m} for the static walking gait and the trotting gait, respectively. The hip height (where we also assume the CoG to be located) is set to \SI{0.38}{m} for all experiments, which corresponds to a height of the base plate of the robot of \SI{0.32}{m}. The gaits' timing parameters (cycle time and duty factor, see Sec.~\ref{sec:gaitselection}) are also hand-tuned. 

We use a cycle time of \SI{2.5}{s} and a duty factor of 86\% for the static walking gait. The trotting gait proves to work best at a cycle time of \SI{0.7}{s} and a duty factor of 70\%. The trotting gait has a very short cycle time because it is only dynamically stable. Furthermore, we had to avoid frequencies that destabilized the motion of the base.

As expected, a significantly higher sinkage was observed with the point feet, which sank up to \SI{120}{mm} into the soil (up to the lowest joint of the parallel linkage). The planar feet sank minimally into the soil (up to the bottom plate of the foot) and always passively adapted to the slope's inclination. The performance notably diverges between the foot designs and the type of gait at inclinations steeper than 20\degree.

During ascent at 25\degree, the planar feet allow the robot to slowly but safely traverse the testbed slope. However, the robot encounters high slippage when using a static gait. Slippage results from the shearing of the uppermost layers of soil that is particularly pronounced as the testbed's inclination approaches the soil's angle of internal friction. Locally, centimeter-scale roughness leads to sections on the slope that might exceed the internal friction angle, making these sections particularly prone to failure through shearing. A faster but more aggressive trotting gait ultimately leads to decreased performance since the soil loosened by the front feet slides onto the hind feet, burying them in the soil. In contrast, the point feet are remarkably stable on a 25\degree~slope when using a trotting gait and static walk, as the increased foot sinkage prevents the soil from shearing. We are confident to say that also steeper inclinations could potentially be ascended with the point feet. However, most of the foot trajectory is executed within the sand as the robot moves forward, imposing the risk of stumbling over unperceived, hidden obstacles, as expected on the Moon and Mars \cite{carrier}.

During descent at 25\degree, the robot equipped with planar feet is "surfing" on the slope substrate, while the static gait keeps the robot stable. In contrast, the point feet considerably sink into the soil, and the resulting traction desynchronizes the robot's static walk, destabilizing the robot and ultimately producing falls. The more balanced trot helps to stabilize the robot with point feet and allows it to descend the inclined testbed, although deep sinkage into the soil is still observed.

\begin{figure}[!tb]
    \centering

    \begin{minipage}{.49\linewidth}
\centering
\subcaptionbox{}{\includegraphics[width=0.47\linewidth]{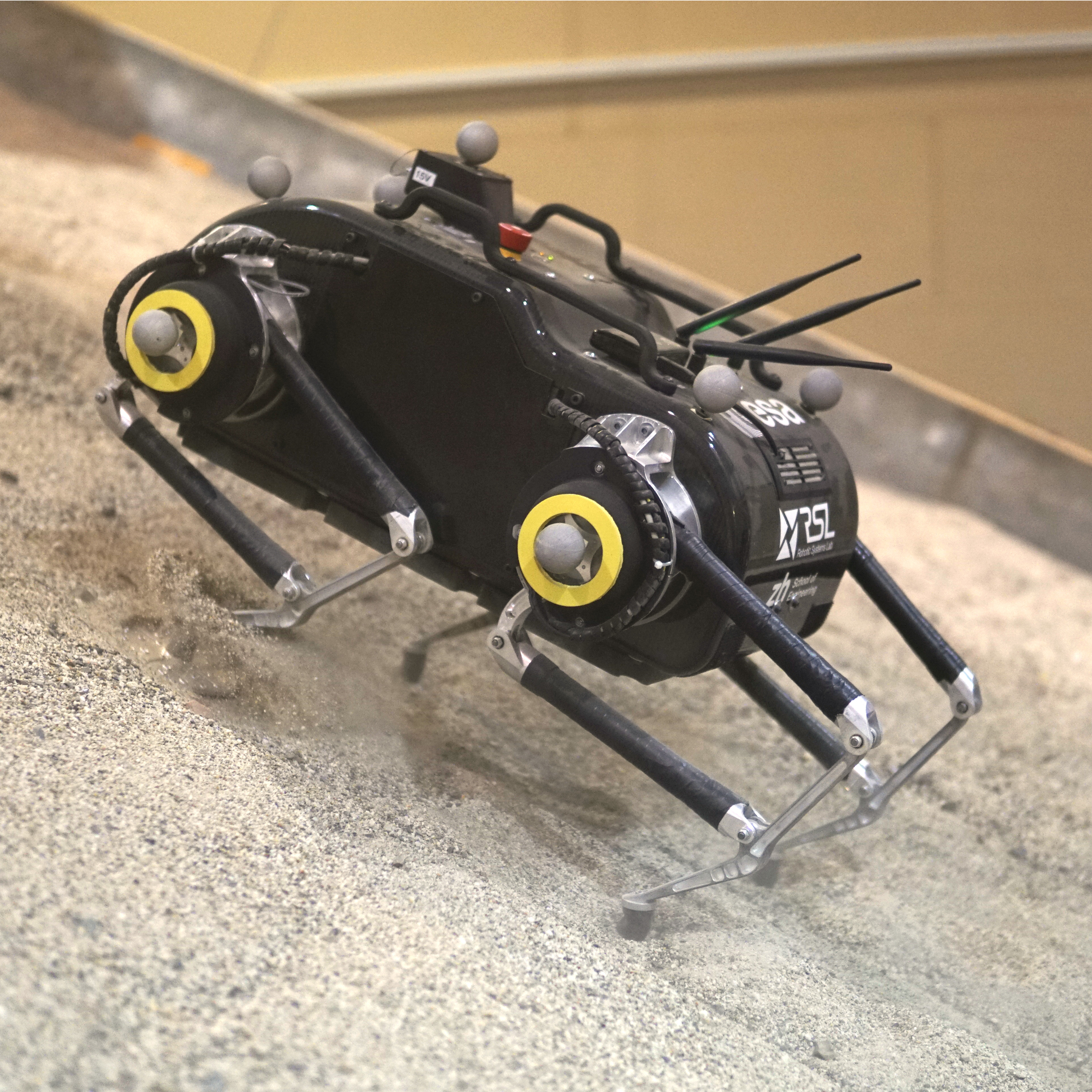}}
\subcaptionbox{}{\includegraphics[width=0.47\linewidth]{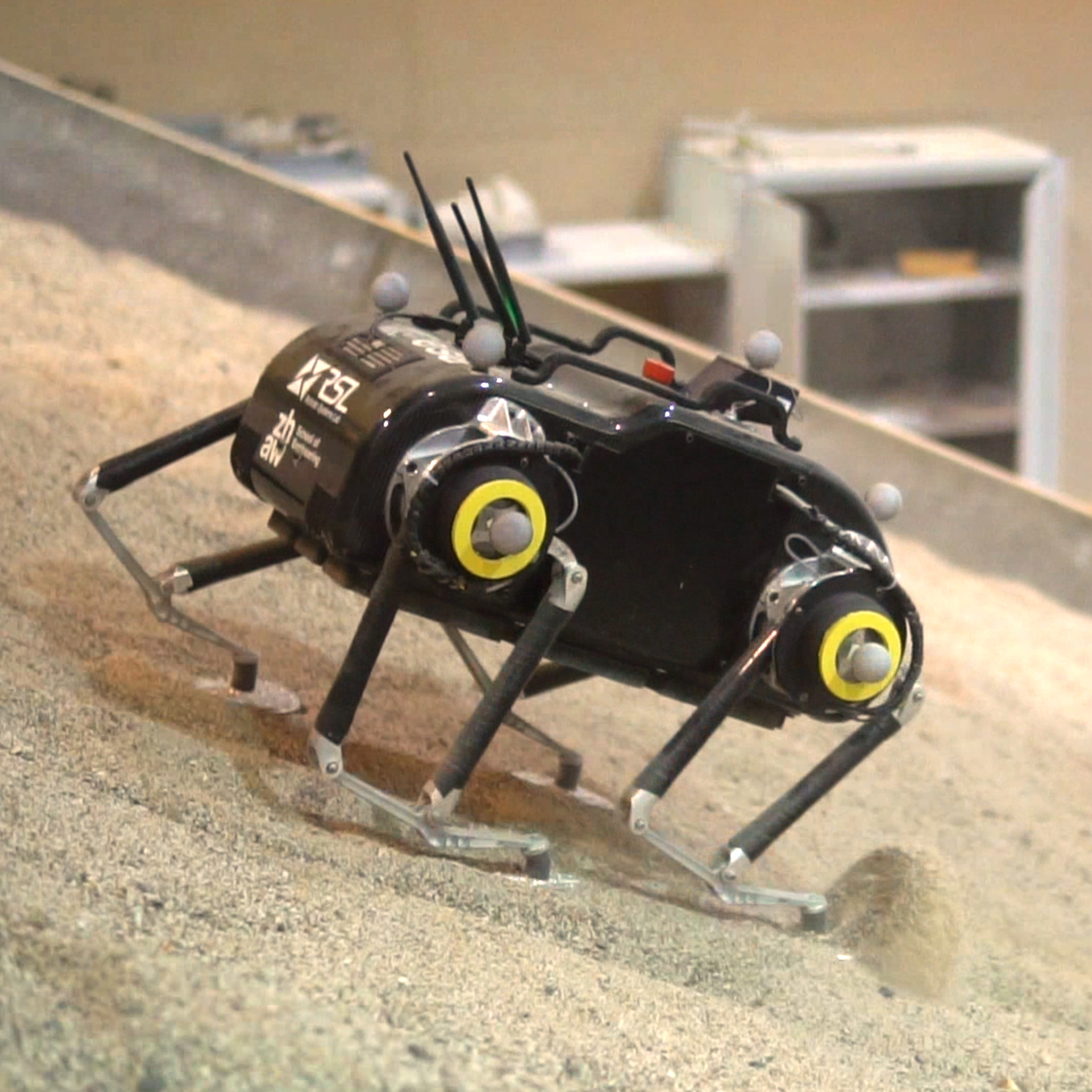}}

\subcaptionbox{}{\includegraphics[width=0.47\linewidth]{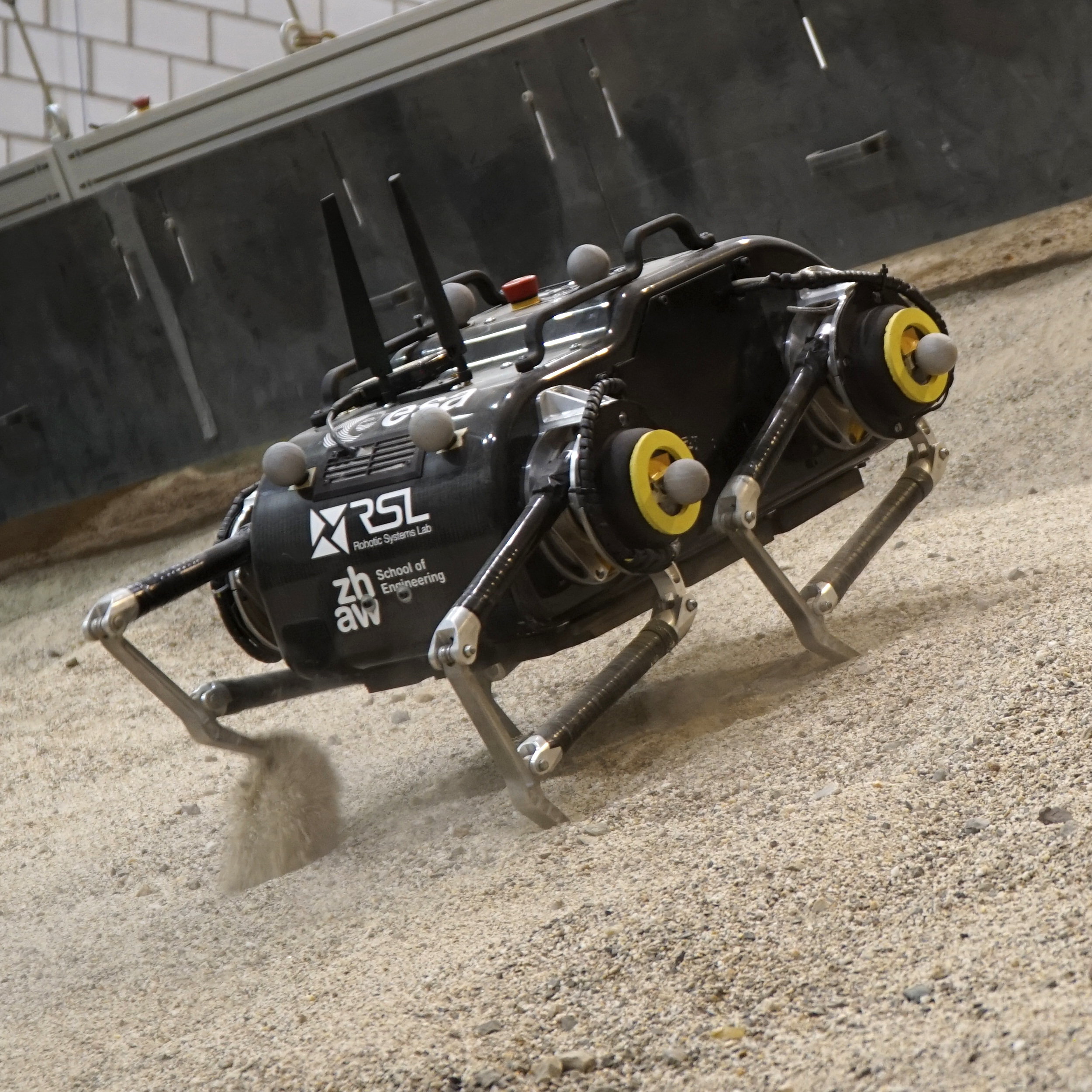}}
\subcaptionbox{}{\includegraphics[width=0.47\linewidth]{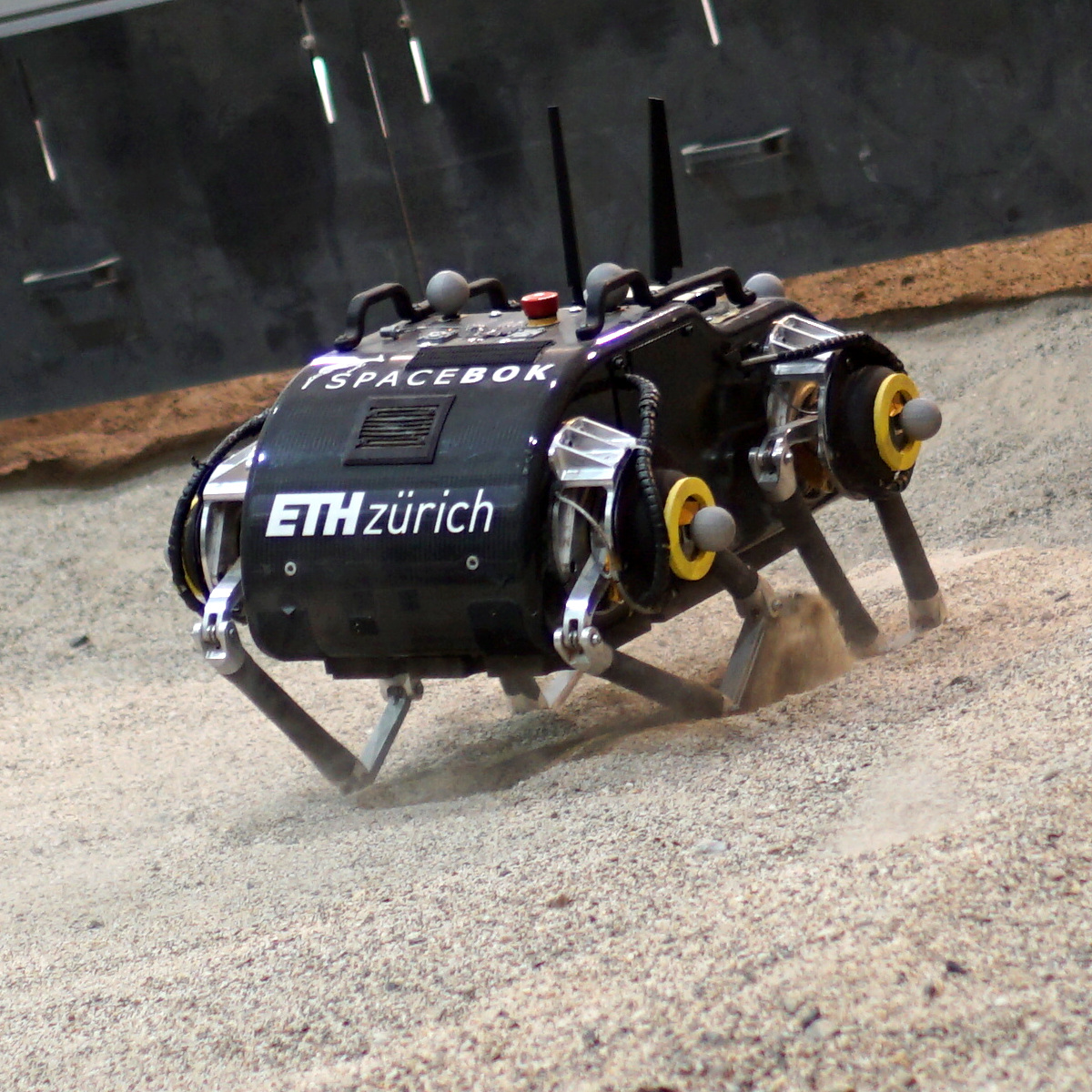}}
    \end{minipage}
        \hfill
            \begin{minipage}{.5\linewidth}
        \centering
        \includegraphics[width=1\linewidth]{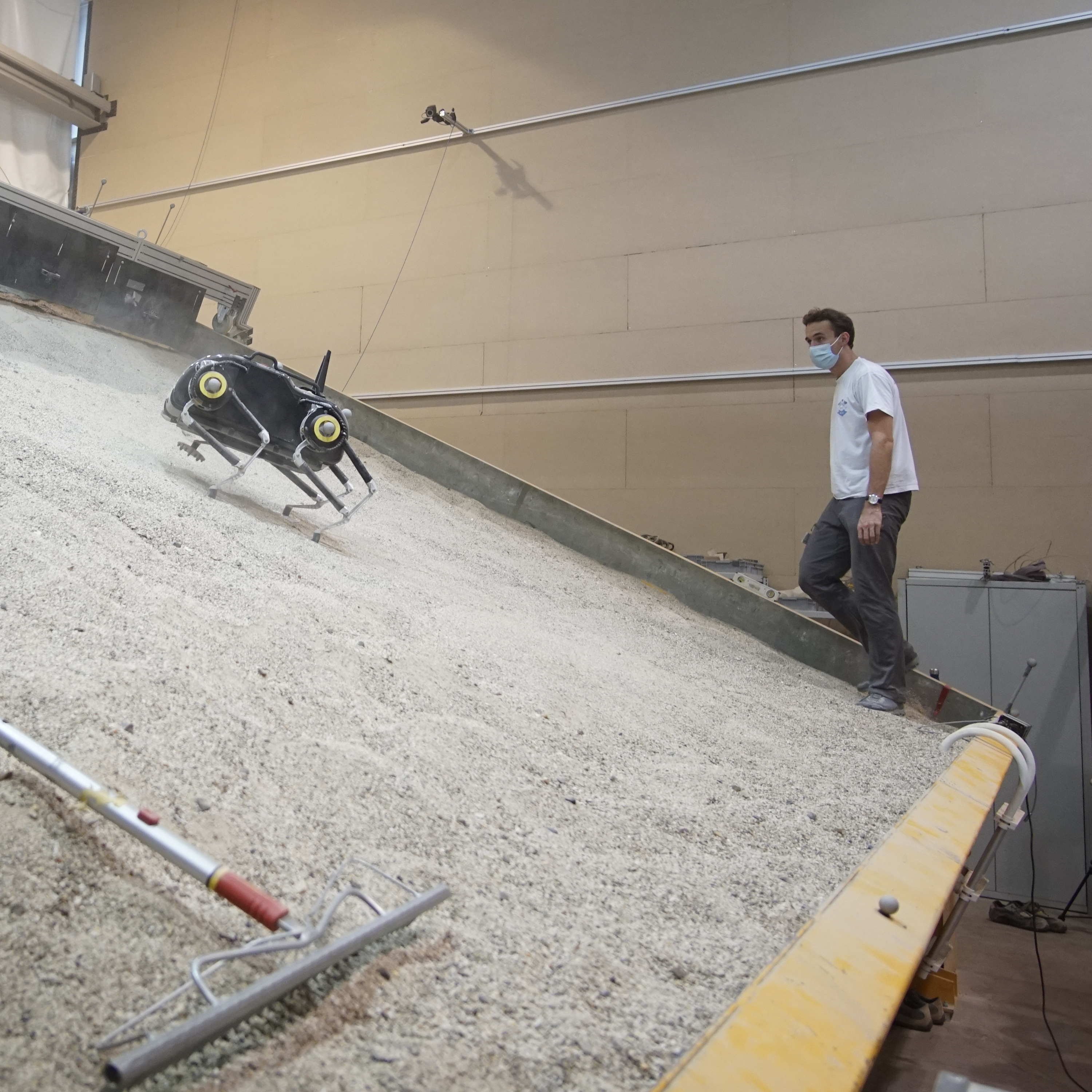}
        \subcaption{}
    \end{minipage}
\caption{Pictures of the static walking experiment on a 25\degree~inclined slope. Depiction (a) and (b) show the robot with planar feet, while (c) and (d) show the robot with point feet. Picture (e) shows the robot with planar feet ascending diagonally with an angle of attack of 45\degree.}
\vspace{-1em}
\end{figure}

\begin{table}[!tb]
\caption{Qualitative evaluation of the slope walking experiment. The pie indicates the stability of the gait for walking. Assessment was made based on observation and number of successful trials.}
\centering
\begin{threeparttable}
\begin{tabular}{|l|l|l|c|c|c|c|c|c|c|}
\hline
\multicolumn{3}{|l|}{\multirow{2}{*}{}}& \multicolumn{7}{c|}{\centering Inclination}\\
\cline{4-10} 
\multicolumn{3}{|c|}{}& 0° & 5° & 10°& 15°& 20°& 22.5°& 25°\\ \hline
\multirow{4}{*}{Static walk}  & \multirow{2}{*}{planar foot} & ascent  & \hspace{0.7em}\pie{360}\hspace{0.7em}       &      \hspace{0.7em}\pie{360}\hspace{0.7em}       &      \hspace{0.7em}\pie{360}\hspace{0.7em}       &      \hspace{0.7em}\pie{360}\hspace{0.7em}      &      \hspace{0.7em}\pie{360}\hspace{0.7em}      &    \hspace{0.7em}\pie{270}\hspace{0.7em}     &      \hspace{0.7em}\pie{180}\hspace{0.7em}      \\ \cline{3-10} 
& & descent & \pie{360} & \pie{360} & \pie{360} & \pie{360} & \pie{360} & \pie{360} & \pie{270} \\ \cline{2-10} 
& \multirow{2}{*}{point foot}  & ascent  & \pie{360} & \pie{360} & \pie{360} & \pie{360} & \pie{360} & \pie{270} & \pie{180} \\ \cline{3-10} 
& & descent & \pie{360} & \pie{360} & \pie{360} & \pie{360} & \pie{180} & \pie{90} & \pie{0} \\ \hline
\multirow{4}{*}{Dynamic walk} & \multirow{2}{*}{planar foot} & ascent  & \pie{360} & \pie{360} & \pie{360} & \pie{360} & \pie{360} & \pie{180} & \pie{90} \\ \cline{3-10} 
& & descent & \pie{360} & \pie{360} & \pie{360} & \pie{360} & \pie{360} & \pie{180} & \pie{90} \\ \cline{2-10} 
& \multirow{2}{*}{point foot}  & ascent  & \pie{360} & \pie{360} & \pie{360} & \pie{360} & \pie{360} & \pie{270} & \pie{270} \\ \cline{3-10} 
& & descent & \pie{360} & \pie{360} & \pie{360} & \pie{360} & \pie{270} & \pie{180} & \pie{180} \\ \hline
\end{tabular}
 \begin{tablenotes}
      \small
    \item  \hbox{\pie{360} = High safety, \pie{0} = Non-traversable}
    \end{tablenotes}
    \end{threeparttable}
\label{tab:qulitative_test_results}
\end{table}

We further analyzed the potential of walking with a non-zero AOA, $\alpha$, since we observed that a frontal 25\degree~climb causes a significant decrease in performance. 

\begin{figure}[!tbh]
	\centering
		\begin{subfigure}[t]{0.200 \textwidth}
	\includegraphics[width=\columnwidth]{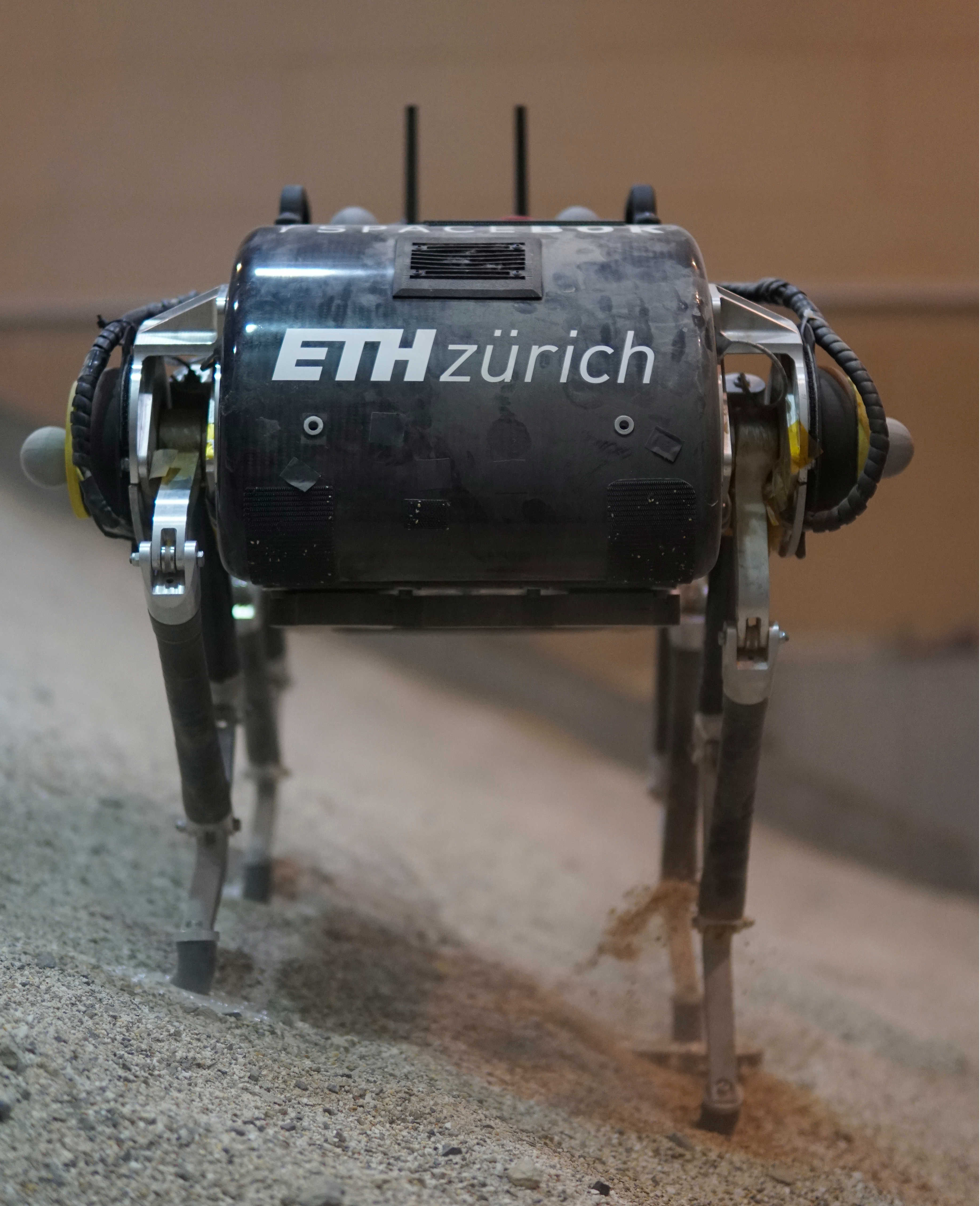}
        \caption{Front view, parallel to the slope with planar feet}
	\end{subfigure}
	\hfill
	\begin{subfigure}[t]{0.283 \textwidth}

	\includegraphics[width=\columnwidth]{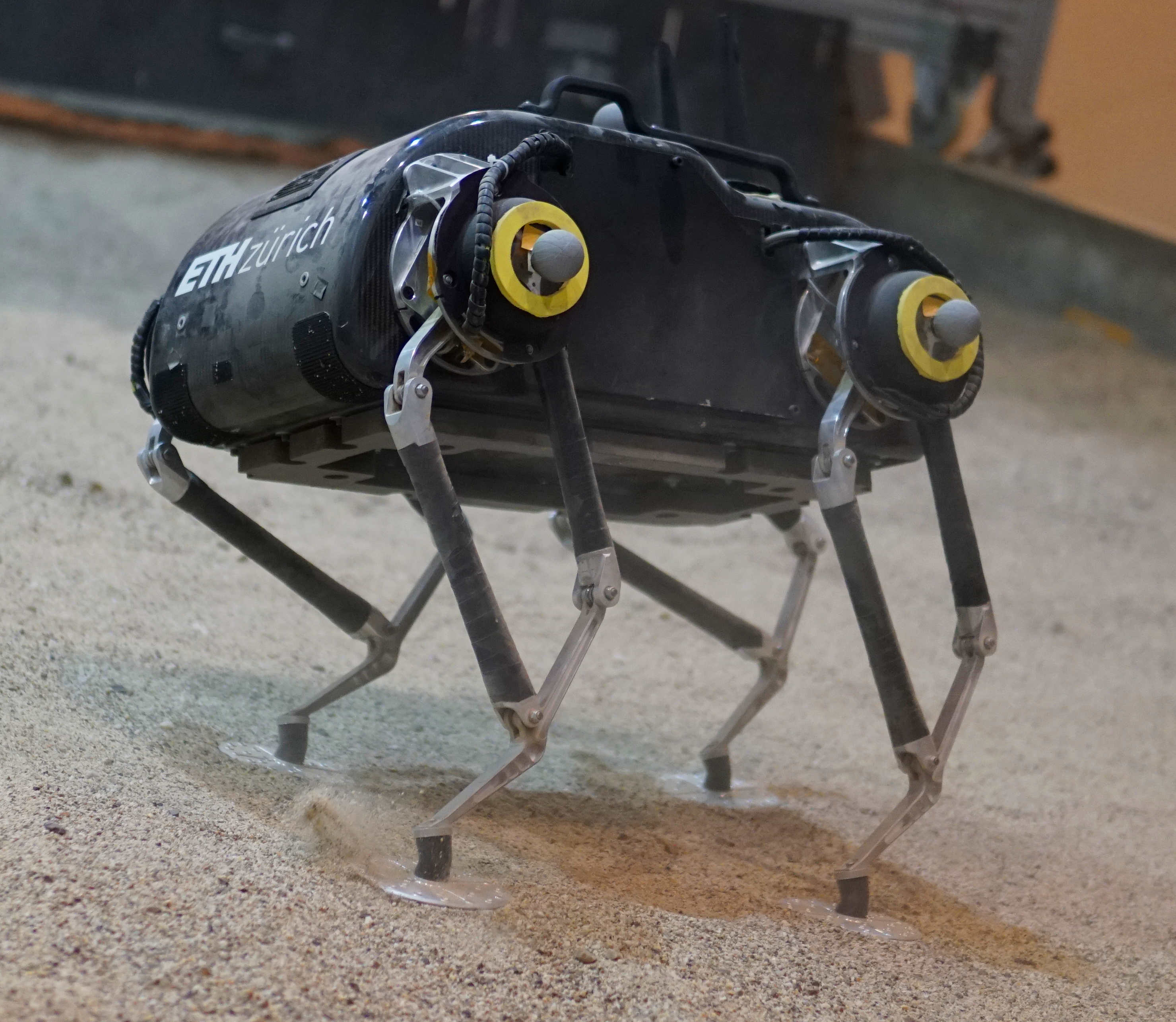}
        \caption{Side view, parallel to the slope with planar feet}
	\end{subfigure}
	\hfill
		\begin{subfigure}[t]{0.200 \textwidth}
				\includegraphics[width=\columnwidth]{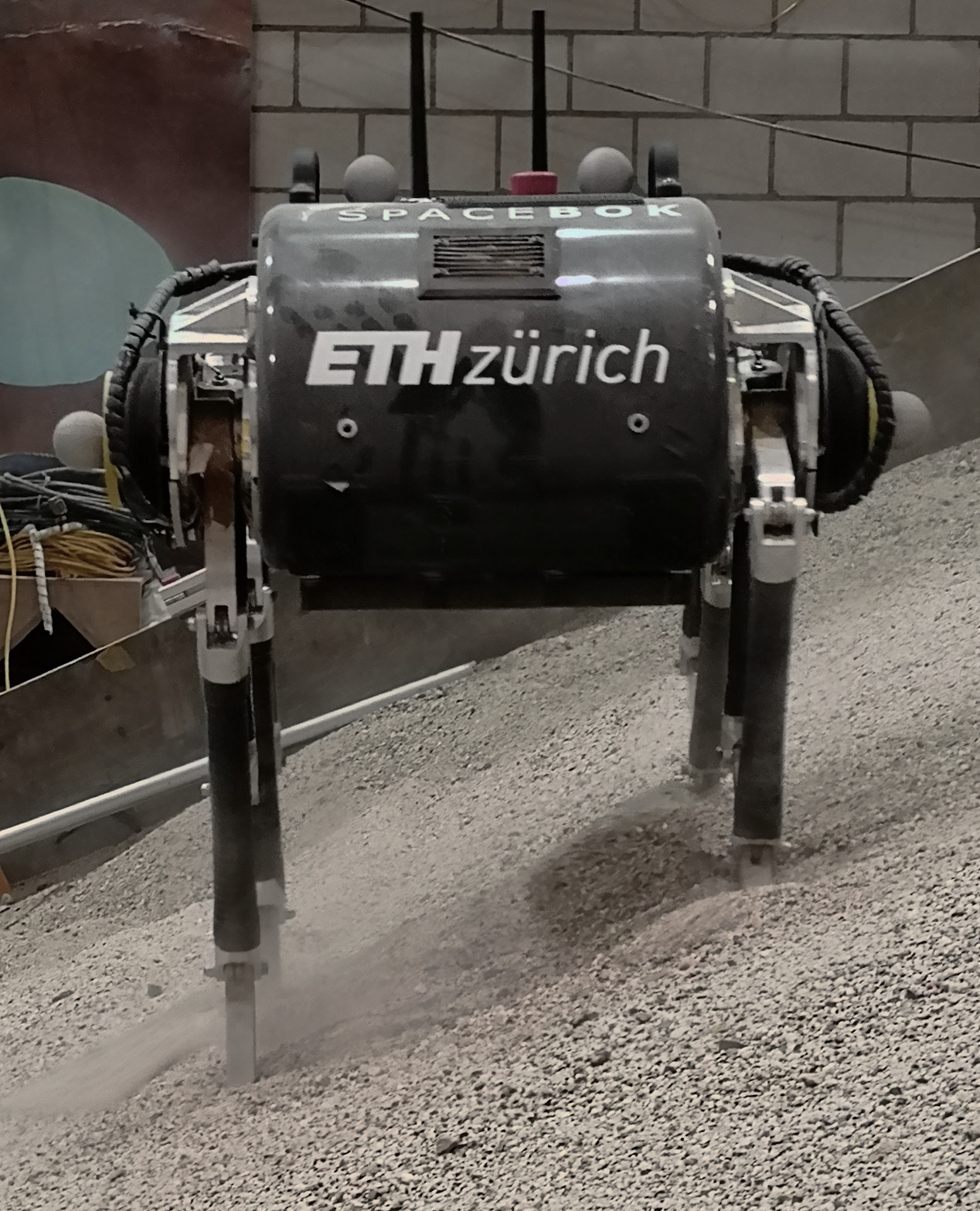}
        \caption{Front view, parallel to the slope with point feet}

	\end{subfigure}
	\hfill
	\begin{subfigure}[t]{0.284 \textwidth}
	\includegraphics[width=\columnwidth]{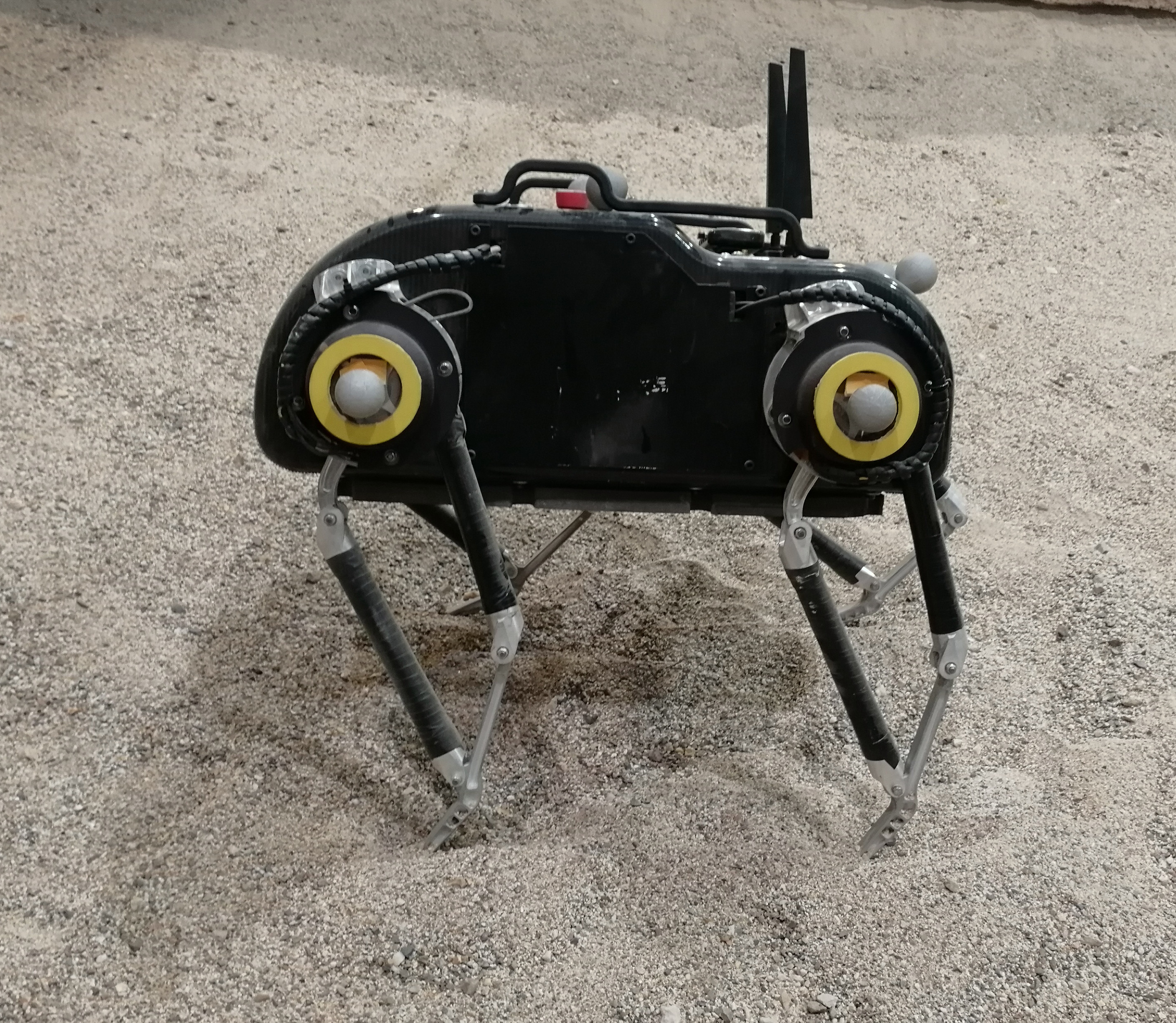}
        \caption{Front view, parallel to the slope with point feet}
	\end{subfigure}
	\caption{Walking parallel to a slope of 25\degree~reveals the advantage of reduced sinkage into the soil. Point feet tend to sink deeper and reach the dangerous singularity configuration on the leg.}
	\label{fig:parallel_slopewalk}
        \vspace{-5pt}     
\end{figure}

With an AOA of below ~43\degree, the encountered inclination in heading direction remains below 20\degree, which showed good performance with all gaits and feet. However, as seen in Figure \ref{fig:parallel_slopewalk}, the large sinkage of the point feet becomes critical when walking in parallel to the slope since the robot's downhill legs are reaching the singularity configuration, which increases the risk of falling. In contrast, the planar feet adapt to the inclination, and no significant sinkage is visible. During the walks, the operator had to steer slightly uphill to counteract side-slip actively.

\begin{figure}[!tb]
\centering
	\begin{subfigure}[t]{0.3 \textwidth}
	\includegraphics[width=\columnwidth]{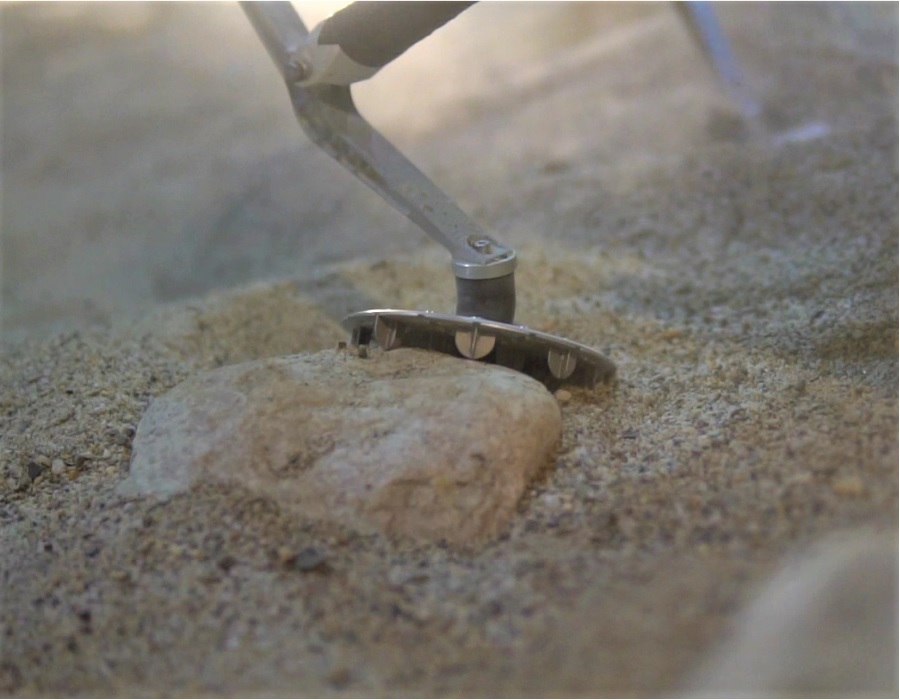}
        \caption{}
	\end{subfigure}
	\begin{subfigure}[t]{0.3 \textwidth}
	\includegraphics[width=\columnwidth]{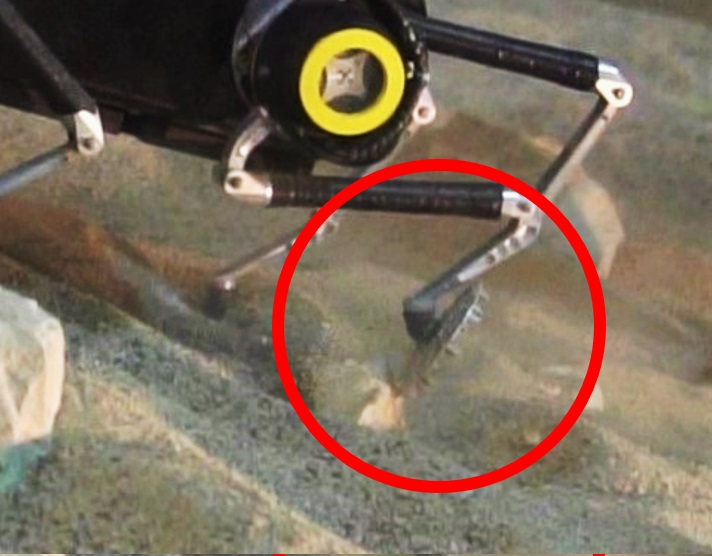}
        \caption{}
	\end{subfigure}
	\caption{Performance of the planar adaptive feet on stones. The feet adapt well when stepping onto a stone (a), but can get stuck in front of a stone (b).}
	\label{fig:foot_rock}
    \vspace{-5pt}     
\end{figure}

Additionally, we conducted preliminary tests on a slope of 20\degree~that has been equipped with obstacles, specifically centimeter-scale cobbles and boulders, to see how the robot performs when transitioning from soft to hard, irregular contacts. While the planar feet adapted well when standing on a rock, they occasionally got stuck underneath an obstacle, as seen in Figure~\ref{fig:foot_rock}.

\subsubsection{Environmental influence}
\begin{figure}[!tb]
\centering
		\begin{subfigure}[t]{0.25 \textwidth}
	\includegraphics[width=\columnwidth]{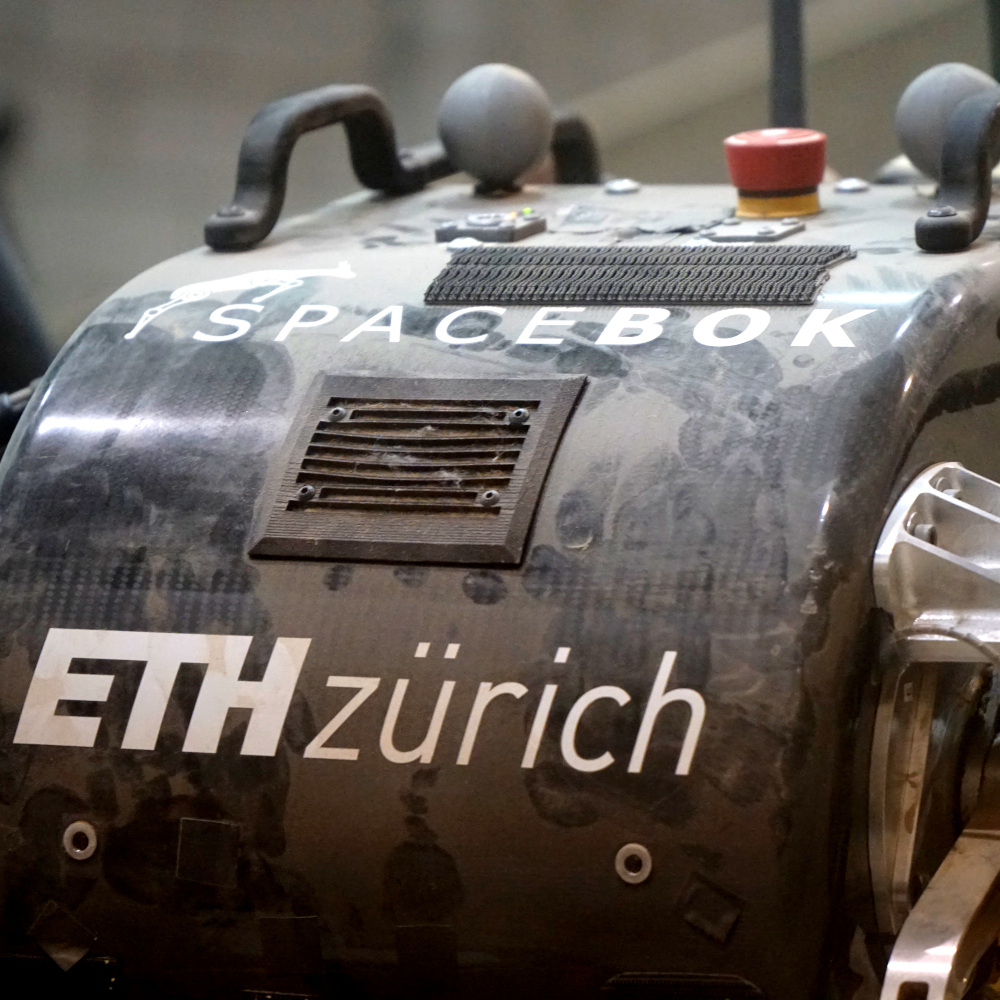}
        \caption{}
	\end{subfigure}
	\begin{subfigure}[t]{0.25 \textwidth}
	\includegraphics[width=\columnwidth]{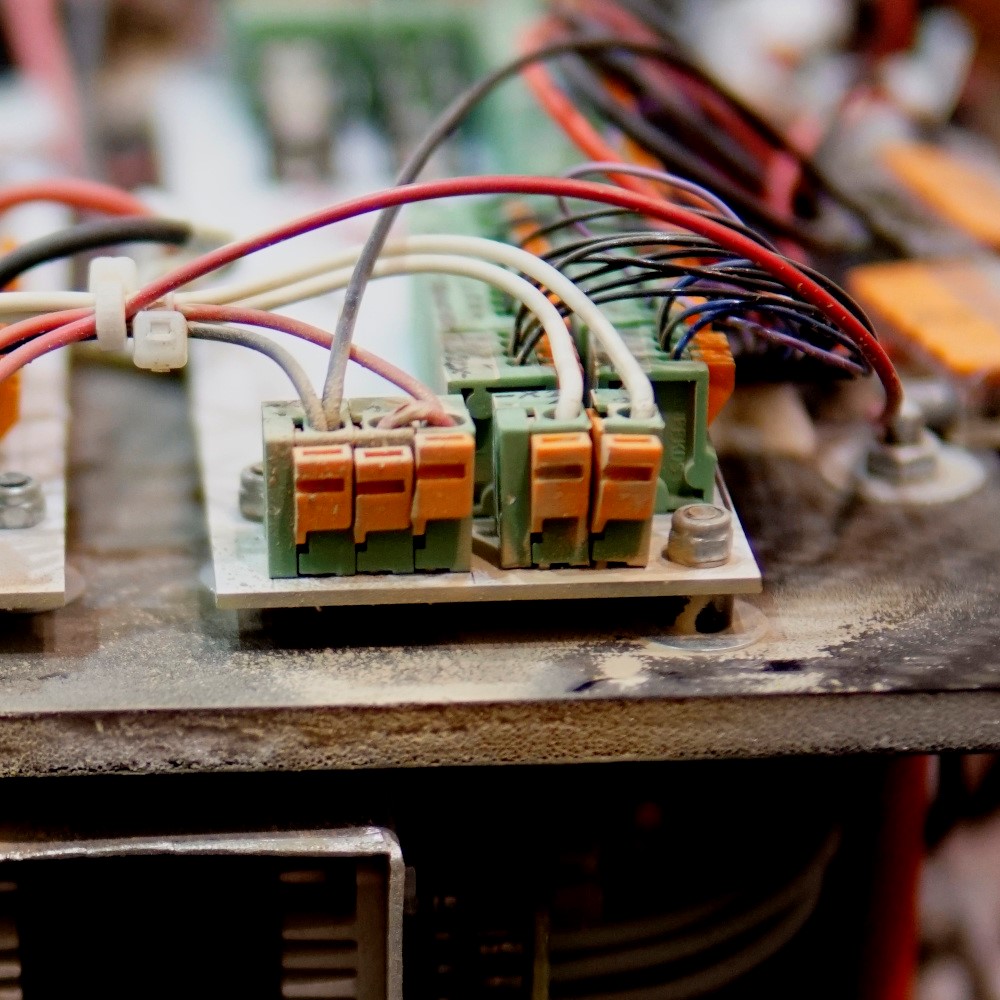}
        \caption{}
	\end{subfigure}
		\begin{subfigure}[t]{0.25 \textwidth}
		\includegraphics[width=\columnwidth]{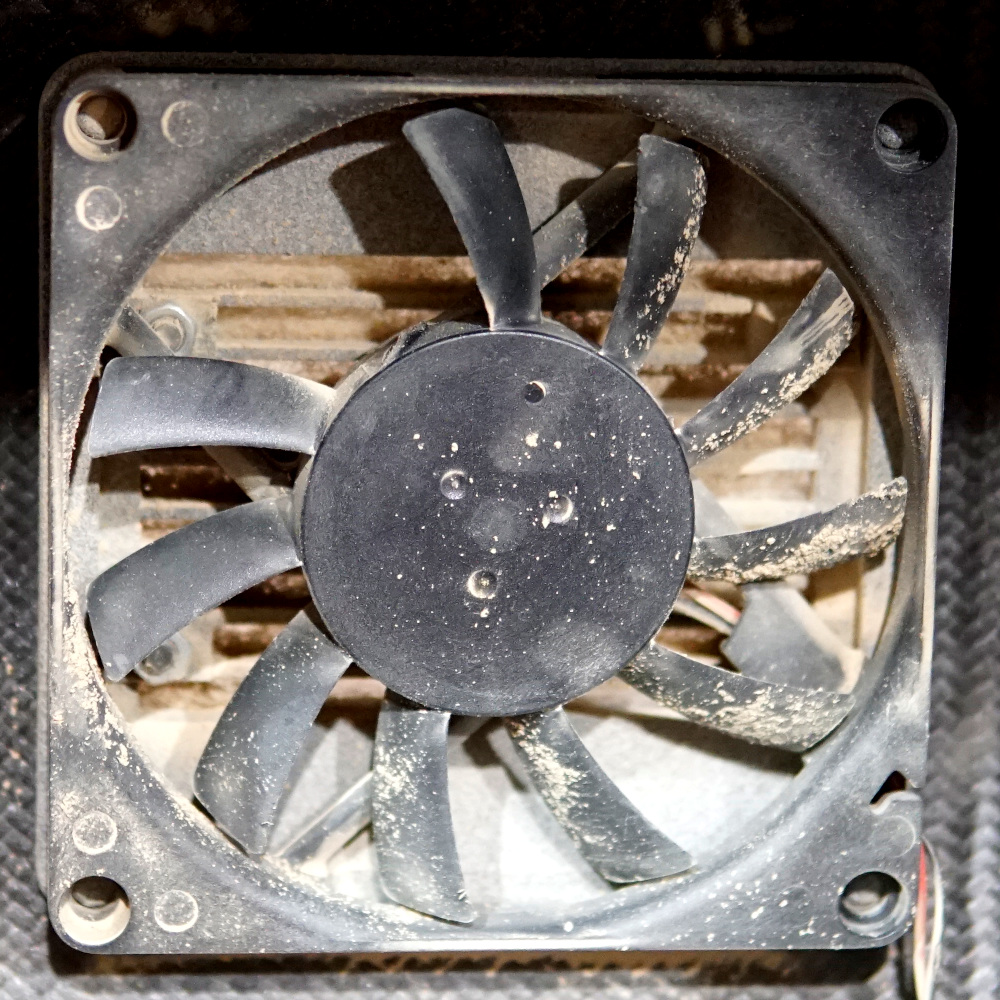}
        \caption{}
	\end{subfigure}
	\caption{Dust accumulation inside the robot after one day of testing. Dust settled on the robot (a), electronics (b) and on the fan and filter (c). Dust is one of the major environmental hazards on the Moon and on Mars \cite{gaiernasa}.}
	\label{fig:dust_in_robot}
        \vspace{-5pt}     
\end{figure}

The frequent soil preparations in combination with the robot operations lead to a large generation of fine dust. While the robot was ingress protected before the field trial, a certain number of small sand/silt particles entered the main body and settled on the electronics (Figure \ref{fig:dust_in_robot}). The subsequent clogging of the filters leads to a reduced cooling performance of the components inside and connected to the main body. Together with the demanding slope-walking maneuvers, the actuator temperature was rising quickly. Consequently, we had to take cool-down breaks towards the end of a test day and thoroughly clean the robot after each day of testing. Given the issues that earlier missions to the Moon and Mars encountered concerning dust (e.g., \cite{gaiernasa}), such as vision obscuration, clogging of mechanisms, and thermal control problems, this observation underlines the importance of effective dust protection measures for a legged space exploration robot.

\subsubsection{Slope dependant velocity and energy consumption}
\label{subsec:ruag_energy_consumption}
We evaluated the energy consumption of the robot in each run. To this end, the electrical power, which is calculated by multiplying the root mean squared (RMS) values of Current $I_{rms}$ and Voltage $V_{rms}$ from the BMS, is normalized with the heading velocity $v_{act}$, calculated from the motion capture system. We subtracted the non-actuation related standby energy consumption of $P_{standby} = \SI{100}{W}$ to only compare the energy consumed for locomotion. The energy per distance value is then calculated as

\begin{equation}
\label{eq:ruag_energy_consumption}
    E = \frac{I_{rms} \cdot V_{rms} - P_{standby}}{v_{act}}.
\end{equation}

\begin{figure}[!tb]
	\centering
		\begin{subfigure}[t]{0.48 \textwidth}
	\includegraphics[width=\columnwidth]{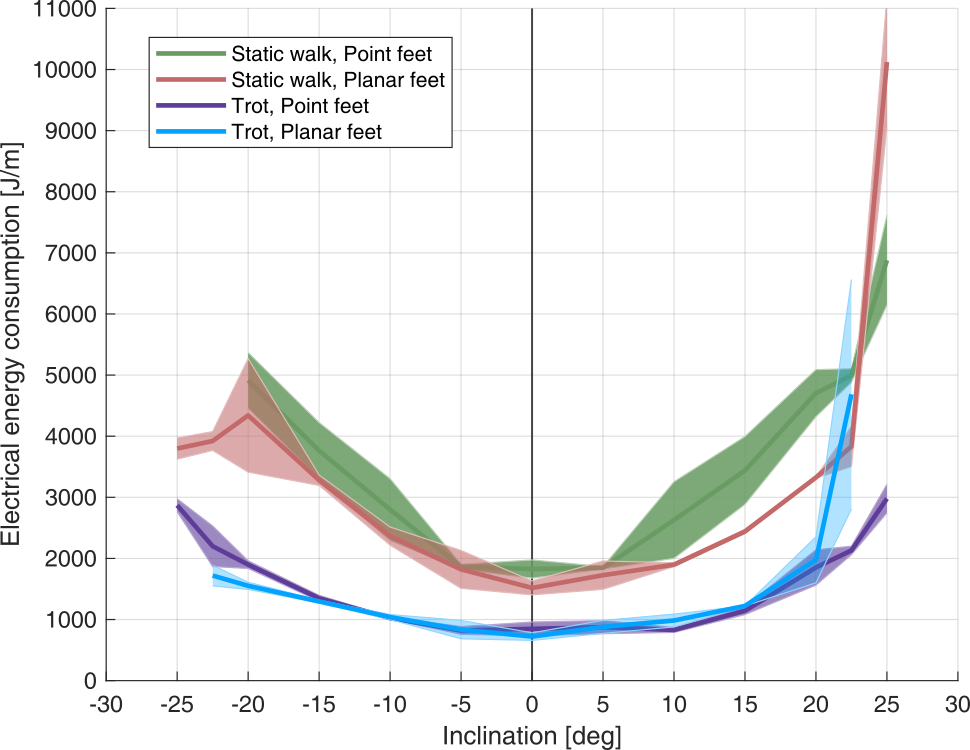}
        \caption{Slope-depending energy consumption }
        \label{fig:ruag_straight_energy}
	\end{subfigure}
	\hfill
	\begin{subfigure}[t]{0.48 \textwidth}
   		\includegraphics[width=\columnwidth]{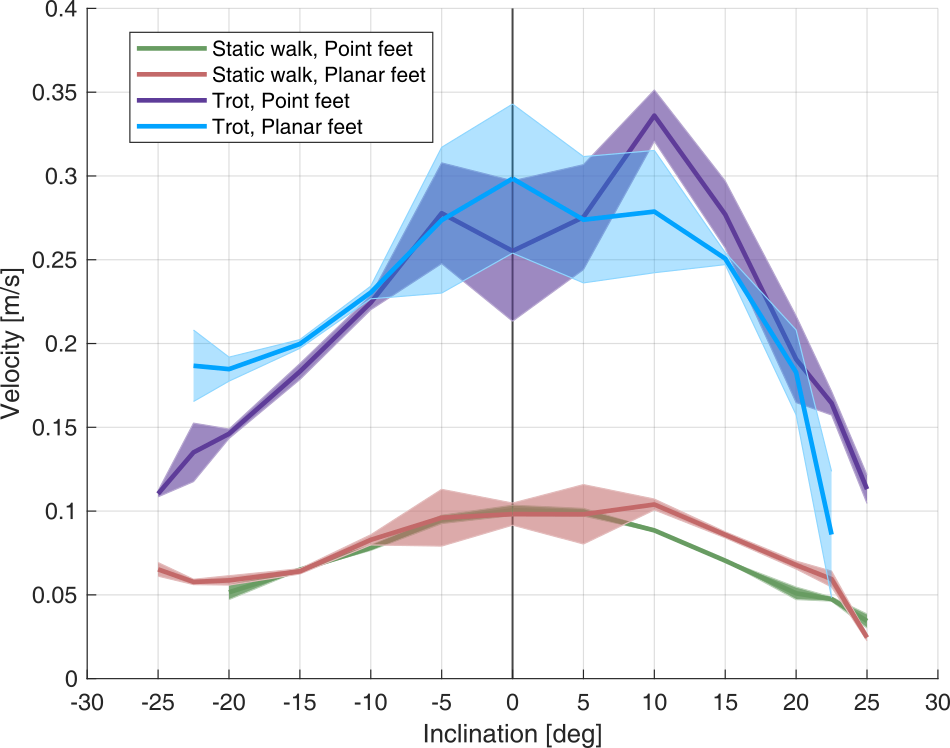}
   		\caption{Slope-depending velocity}
   		\label{fig:ruag_straight_velocity}
	\end{subfigure}
		\begin{subfigure}[t]{0.48 \textwidth}
	\includegraphics[width=\columnwidth]{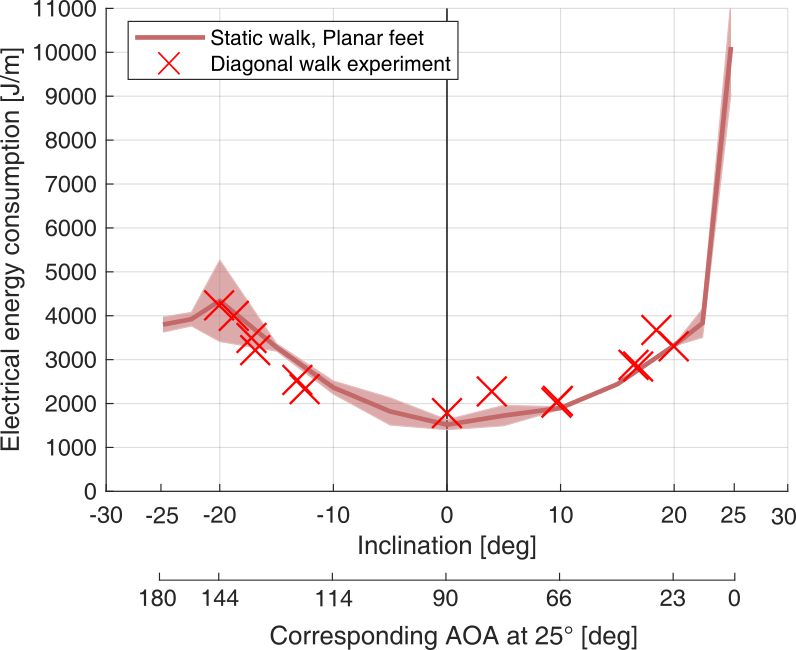}
        \caption{Diagonal walk with planar feet}
	\label{fig:ruag_diag_planar}
	\end{subfigure}
	\hfill
	\begin{subfigure}[t]{0.48 \textwidth}
   		\includegraphics[width=\columnwidth]{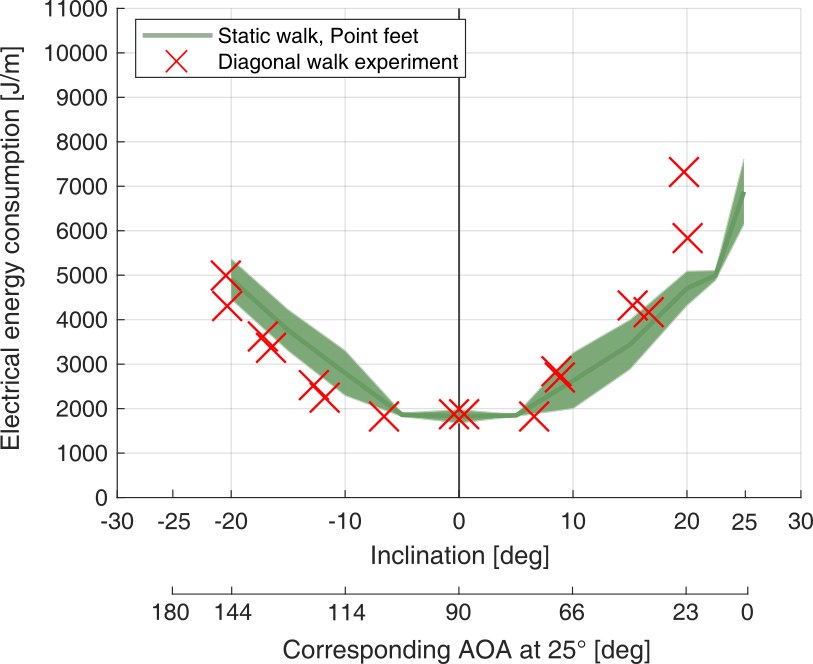}
   		\caption{Diagonal walk with point feet}  			\label{fig:ruag_diag_point}
	\end{subfigure}	
	\caption{Experimentally determined average values for energy consumption and achievable velocity in relation to different locomotion gaits and feet designs. The standard deviation is indicated as shaded regions in the plot.}
	\label{fig:ruag_energy}
\end{figure}

\begin{figure}[!tb]
\centering
   \includegraphics[scale=0.38]{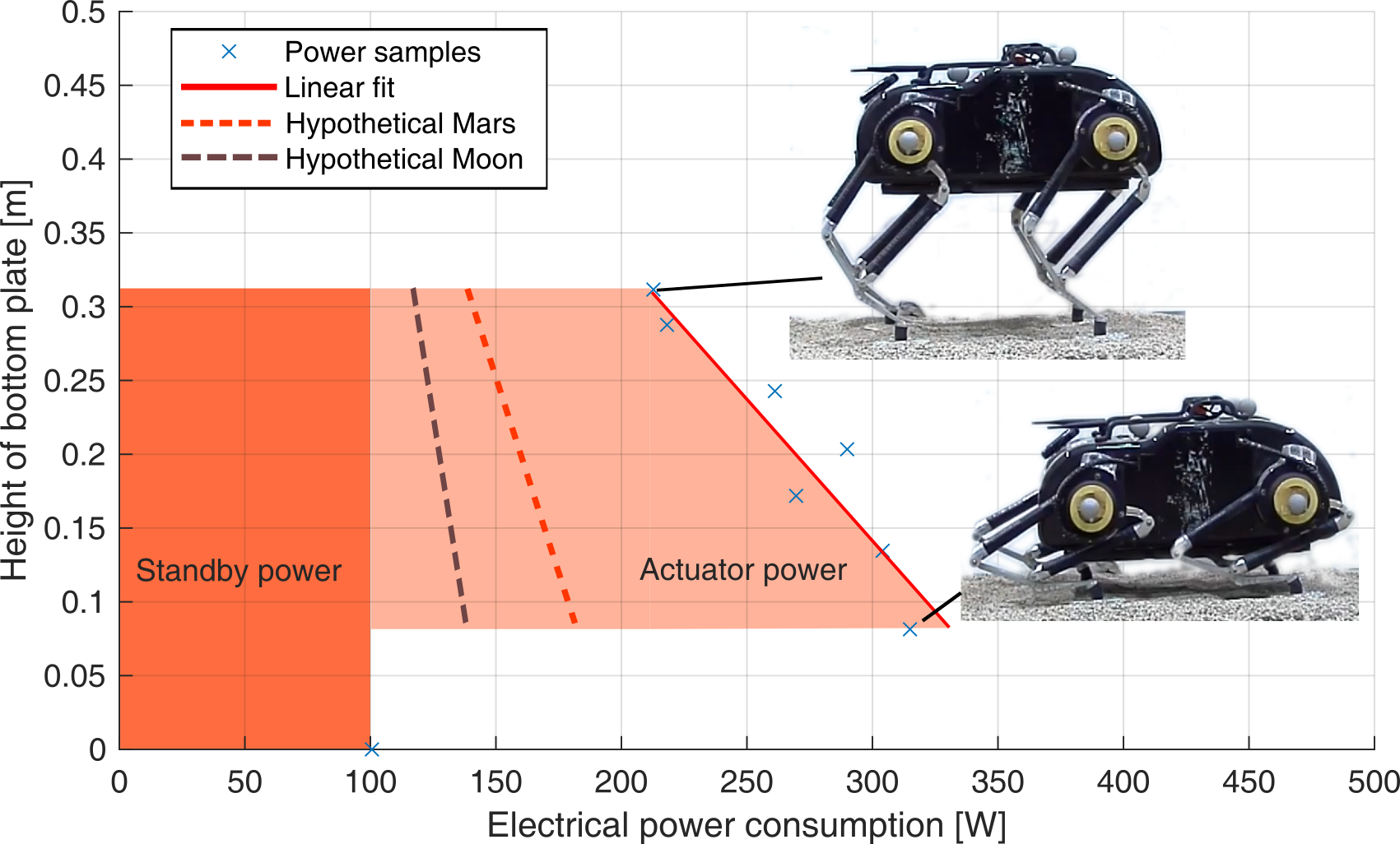}
\caption{The average power consumption of the robot in relation to base height. The robot consumes \SI{100}{W} without powered actuators and power consumption increases from \SI{220}{W} (extended legs) to \SI{325}{W} (crouching).}
    \label{Fig:standing_power}
\end{figure}

As seen in Figure \ref{fig:ruag_straight_energy}, the dynamic gait is more than twice as efficient as the static walking gait for both foot designs. This correlates to the speed, which is twice as fast during the trotting motion. While the energy consumption rises almost linearly up to around 20\degree, a notable increase in energy consumption is visible for steeper inclines. The energy increase is evident with the planar feet, where the consumption almost triples. This correlates to the observed slip and thus reduced velocity on slopes close to the soil's internal friction angle. Interestingly, the point feet seem to have a slightly higher energy demand with the static gait at slopes below 20\degree~compared to the planar feet, which might be related to energy loss due to work performed on soil deformation and thus reduced speed. The velocity and time depending energy consumption are explained by the robot's average power to suspend its body weight. As seen in Figure \ref{Fig:standing_power}, a significant contribution to energy consumption is the suspension of the robot's weight by the actuators requiring about 55-70\% of the robot's total energy budget.

This requires a trade-off with regard to the system requirements. While lowering the robot's COG decreases the risk of falling, it significantly increases the power consumption. The actuator power required when standing with stretched legs compared to a fully crouched position is twice as high, and taking into account the standby power, the system's overall power demand rises by 50\%. However, in the reduced gravity setting of the Moon or Mars, the power consumption for suspending the robot's weight reduces to one-sixth or one-third, respectively, thus becomes more attractive.

A strategy to efficiently overcome steep inclinations is to walk in curves with a non-zero AOA. Figures \ref{fig:ruag_diag_planar} and \ref{fig:ruag_diag_point} show the energy consumption of the robot when walking in diagonal paths on the 25\degree~slope. An AOA of 0\degree~corresponds hereby to a straight, head-on climb, 90\degree~to a parallel walk, and 180\degree~to a straight descent. Overlaying the energy data from the diagonal walk experiment with the previously sampled data from direct ascent and descents shows a similar trend, which indicates that energy losses due to side slip have a small influence on the performance. Higher energy consumption is noted with the point feet at an AOA of 23\degree, which corresponds to a 20\degree~slope in the heading direction. The effect is likely to result from the active steering of the robot operator to keep the heading, thus forcing the robot to slip purposefully and reduce its velocity. In general, we consider the inclination in the heading direction to be a good first-order approximation for energy consumption and neglect energy losses due to the problem's bidirectionality.

\begin{figure}[!tb]
\centering
		\begin{subfigure}[t]{0.4\textwidth}
	\includegraphics[width=\columnwidth]{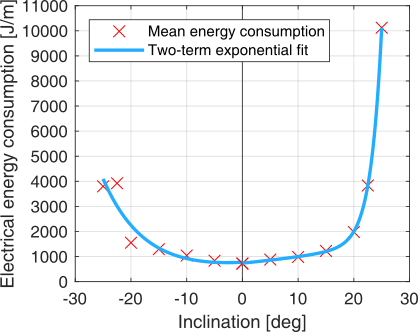}
        \caption{Planar foot}
	\end{subfigure}
	\begin{subfigure}[t]{0.4 \textwidth}
	\includegraphics[width=\columnwidth]{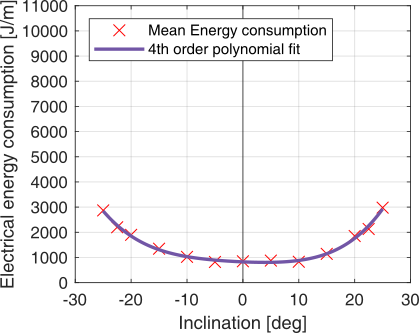}
        \caption{Point foot}
	\end{subfigure}
	\caption{Mean experimental data and function fit for slope-depending energy consumption of planar and point feet.}
	\label{fig:planner_energyfit}
        \vspace{-5pt}     
\end{figure}

Accordingly, we derived slope-depending energy functions (Figure \ref{fig:planner_energyfit}). For the planar feet, we use the energy consumption of a trotting gait up to 20\degree, and switch to a static gait at steeper inclinations. Two (ascent and descent), two-term exponential functions of the form
\begin{equation}
    E_{planar}=ae^{bx}+ce^{dx}
\end{equation}
were fitted and evaluated. The respective parameters are
\begin{align*}
    & [-25\degree \hdots 0\degree]: \{a=139.3, b=-0.1333, c=634.7, d=0.0479\} \\
    & [0\degree \hdots 25\degree]: \{a=737.7, b=0.02979, c=0.0233, d=0.5126\} .
\end{align*}
For the point feet, we fitted a single 4th-order polynomial of the form
\begin{align}
    &E_{point} = p_1 \cdot x^4 + p_2 \cdot x^3 + p_3 \cdot x^2 + p_4 \cdot x + p_5 \\
    \text{with} \qquad
    &[-25\degree \hdots 25\degree]: \{p_1=0.003763, p_2=0.01768, p_3=0.9459, p_4=-9.302, p_5=825\} . \nonumber
\end{align}
In both functions $x$ represents the encountered slope in degrees. The velocity functions are derived analogously. 
As observed in the experiments, degradation in performance occurs when facing slopes with more than 20\degree~inclination, which should be taken into account at the planning stage. With those energy functions, it is possible to derive the optimal angle of attack for a given slope. We solve 
\begin{align} 
\label{eq:analytical_efficiency}
E_{i,tot} &= E_{i}(\gamma,\beta ) l(d,\alpha ),
\end{align}
where $E_{i}(\gamma,\beta )$ is the energy consumption per meter in (diagonal) heading direction, which is based on the overall slope inclination $\gamma$ and the angle of attack $\alpha$, and the increased distance to the goal $l(d,\alpha )$, which is a function of the shortest distance $d$ and the angle of attack $\alpha$. Solving for the local Minima of the energy functions, we derive an optimal AOA for ascending a 25\degree~slope to be 54\degree~for the point foot and 50\degree~for the planar foot.

\section{Energy aware path-planning on slopes}\label{sec:mars_path_strategies}

\subsection{Path-planning strategy}
We used the results from the slope-walking experiment to find energy-efficient paths through an exemplary crater on Mars. While climbing steep slopes in direct ascent is possible and eventually unavoidable, the actual time spent in such terrain might be relatively small compared to the overall mission. Additionally, we want to compare the energy consumption of the two feet. Thus, we planned a traverse over a large map (width and height of multiple hundreds of meters). In our case, the path-planning problem can be simplified to a 2D problem since the height, and the resulting cost is given at every position by the elevation map. Due to the two DOF legs of the robot, a limited turning radius of about \SI{2.5}{m} has to be taken into account, which makes it necessary to consider the yaw orientation during planning. Thus, the robot's state is fully described by $(x,y,\theta) \in SE(2)$. In our approach, we decided to use a sampling-based planner over a graph- or grid-based method. Usually, sampling-based planners are slower to converge to an optimal solution, but because we assume that path-planning can be done offline before starting a mission, the runtime is not a limiting factor. Additionally, collision checking and explicitly considering the kinematic constraint of \textit{SpaceBok} are straightforward to implement.

Specifically, we implemented the RRT* algorithm, which is guaranteed to converge towards an optimal solution over time \cite{frazzoli2011}. Additionally, we used Dubin's curves to constrain the motion between two sampled states to feasible paths. Dubin's curves are motion primitives consisting of three segments where each can be a straight line or an arc segment \cite{dubins1957}. This motion primitive only considers forward motions, unlike others, such as the more general Reeds-Shepp curves, which also cover backward motions \cite{reeds1990}. In our case, we consider a simple scenario in which the robot would mostly walk forward, thus requiring a minimal sensor setup. 

As a cost function in RRT*, we used the fitted functions for the energy consumption determined based on the slope walking experiments. The total energy consumption is then calculated by integrating the energy per meter along the path. For each state, the orientation of \textit{SpaceBok} is taken into account to calculate the slope in the robot's body frame correctly. As an approximation, we only consider the slope along the robot's heading direction, as discussed in Section \ref{subsec:ruag_energy_consumption}. The global path-planning algorithm has been implemented using the Open Motion Planning Library (OMPL) \cite{ompl}. 

\subsection{Scenario description}
We used a HiRISE-derived digital terrain model from a region called "Xanthe Terra" \footnote{\url{https://www.uahirise.org/dtm/dtm.php?ID=ESP_017555_1875} (\text{DTEEC\_017555\_1875\_018544\_1875\_A01.IMG})} \cite{hirise2007}. We selected an area of \SI{400}{m} x \SI{350}{m} that contains parts of a small-scale impact crater with a crater diameter of about \SI{400}{m} and a depth of \SI{70}{m}. The map data is provided at a resolution of \SI{1}{m}. The selected section of the map can be seen in Figure \ref{fig:spacebok_path_planner_data}. To test our global path planner, we use a scenario that is close to a real-world planetary mission where \textit{SpaceBok} has to enter a crater, possibly inspect relevant locations on the crater floor, and then exit the crater again via a relatively steep slope to continue its mission or return samples. \\

In our scenario, we did not incorporate knowledge about the traversal of local irregularities, such as rock abundance, or CFA (Cumulative Fractional Area), which would require further knowledge about the robot's performance on such terrain.

\subsection{Path-planning results}\label{sec:path_planning_results}
\begin{figure}[!bt]
\centering
\begin{subfigure}[t]{0.4 \textwidth}
	\includegraphics[width=\columnwidth]{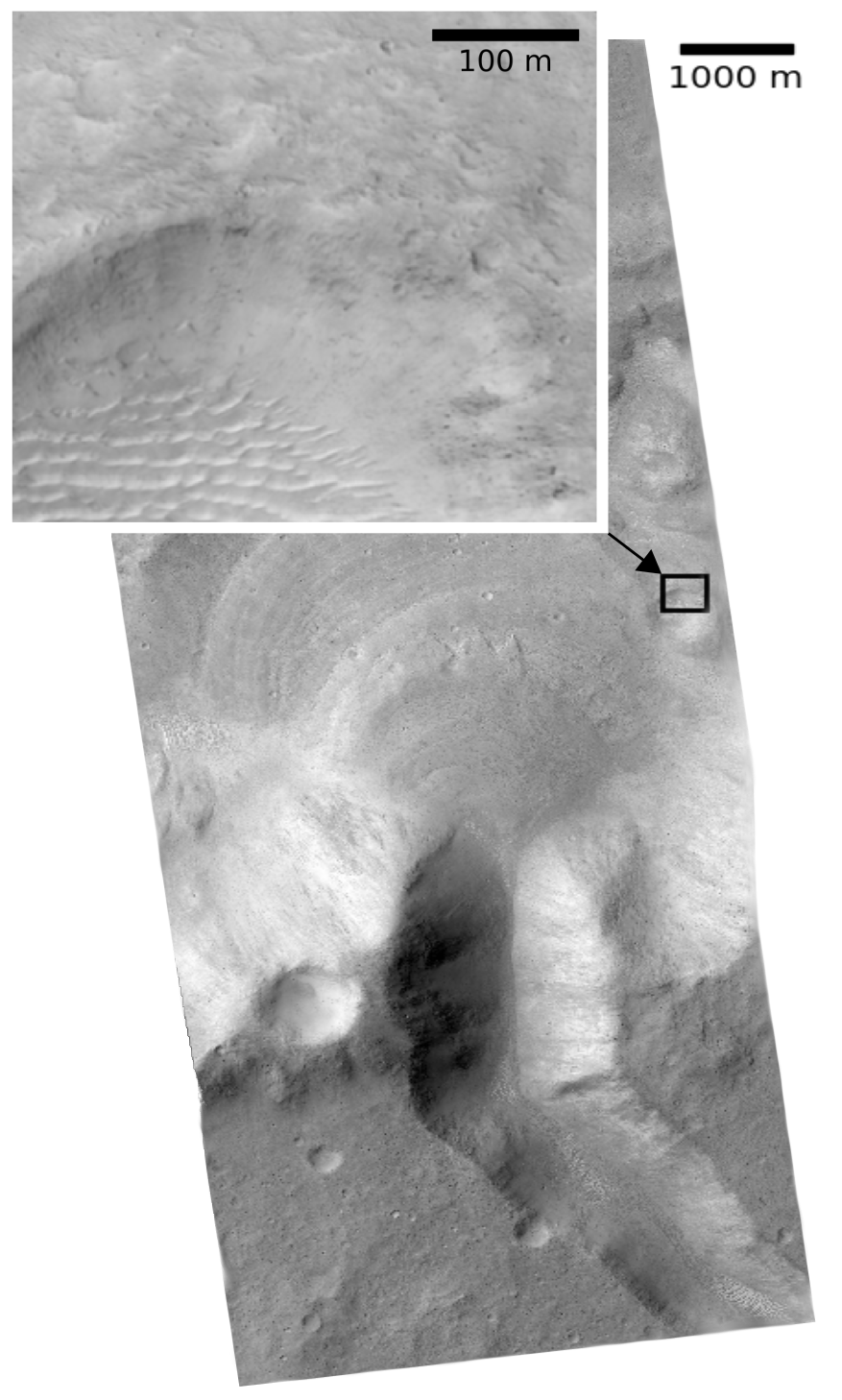}
        \caption{Orthoimage (HiRISE) showing the \newline location of the crater used for planning \newline (NASA/JPL/University of Arizona/USGS)}
	\end{subfigure}
	\begin{subfigure}[t]{0.625 \textwidth}
	\includegraphics[width=\columnwidth]{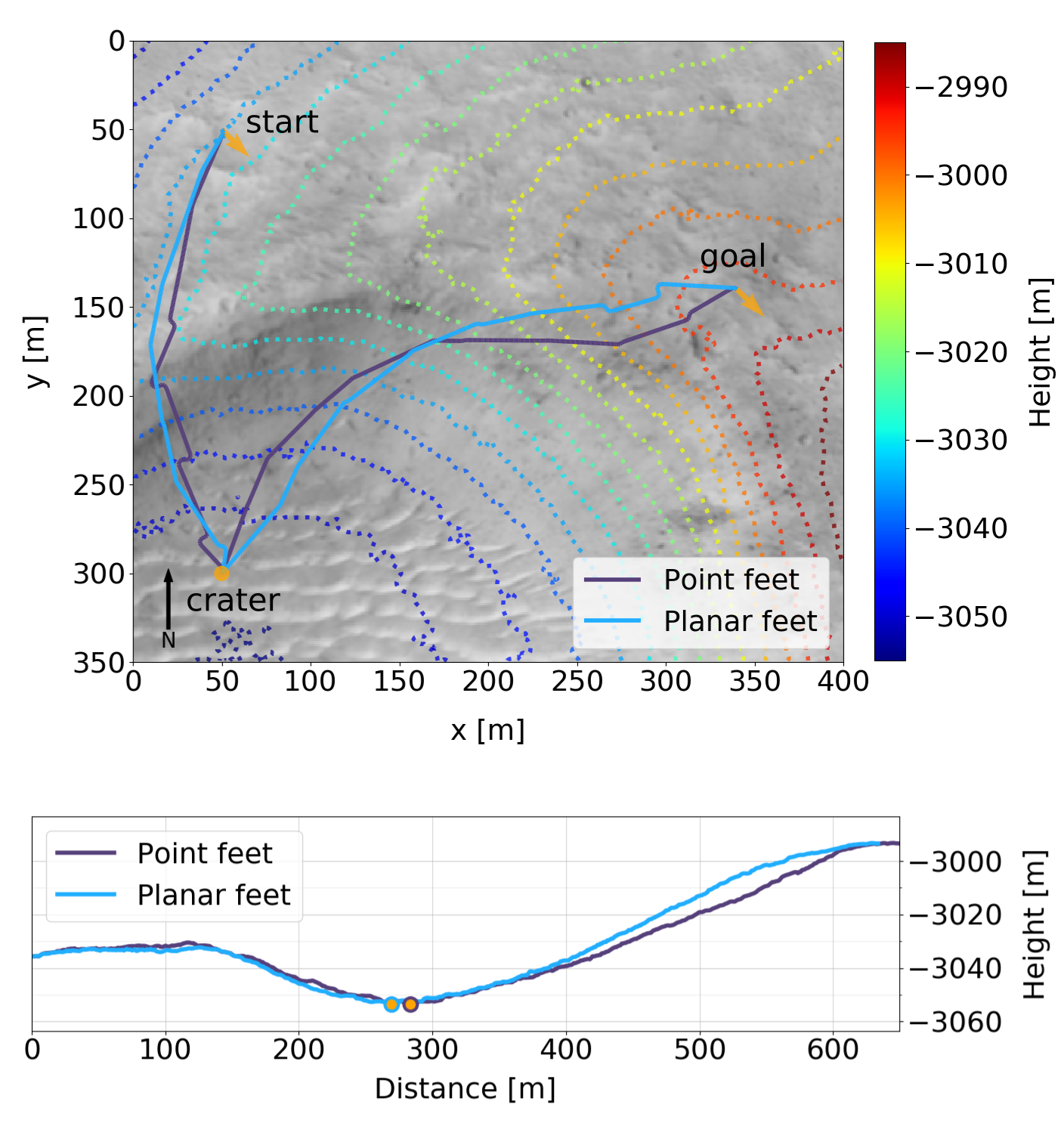}
        \caption{Top-down view and height profile of planned paths. The orange \newline circle marks the boundary between descent and ascent path.}
	\end{subfigure}
	\caption{Crater location and planning results for a traverse on Mars.}
    \label{fig:spacebok_path_planner_data}
\end{figure}

We executed the path planner with two cost functions; the one obtained for the point feet and the one obtained for the planar feet. We had to split the path into two segments to find a path for our scenario since the RRT* algorithm does not support intermediate goals. The first segment starts at the crater rim and ends within the crater. We call this scenario \textit{descent}. The second scenario starts at the crater basin and ends on the crater rim at a different location. We call this scenario \textit{ascent}. The planned paths, including the intermediate goal in the crater, can be seen in Figure \ref{fig:spacebok_path_planner_data}. 

Our path planner found feasible paths for both foot designs that allow \textit{SpaceBok} to accomplish the mission. Especially during ascent, it can be seen that the path is not a straight connection between the crater and the goal. The optimal path makes a slight detour to reduce the angle of attack to the main slope direction in the region where the slope is steep. The histograms of the slope evaluated at along the path in Table \ref{tab:spacebok_path_planner_quantitative_results} confirm this behavior. The resulting paths avoid extreme slopes (above \SI{20}{\degree}) where the energy consumption grows exponentially and where it is risky to walk. 

The quantitative results in Table \ref{tab:spacebok_path_planner_quantitative_results} agree very well with the experimental data in Figure \ref{fig:planner_energyfit}. The planar feet are more energy-efficient when walking downhill, and the point feet are slightly more energy-efficient when walking uphill. Overall, the total energy consumption using different feet was identical with \SI{604}{kJ} for \SI{650}{m} with the point foot, and \SI{602}{kJ} for \SI{633}{m} with the planar foot. We note that those energy values represent the energy consumption by the actuators under Earth-gravity. Since the energy is mostly consumed by suspending the robot's mass, we assume that deployment on Mars would reduce the energy consumption to roughly one-third of the calculated value. While the aforementioned values give an indication of what Energy consumption is to be expected, the specific gait parameters might vary on Mars due to the change in gravity. However, we consider this influence rather small compared to the energy expenditure for suspending the weight of the robot.

We used the maximal velocity per slope to evaluate the resulting paths further as determined in our experiments. The results show that the point feet are marginally faster compared to the planar feet when walking up a slope and at a similar speed when walking down a slope. 
Overall, the difference between the energy consumption of both foot types is minimal since both try to minimize the time spent on slopes above \SI{20}{\degree}, where the main differences occur. This highlights that cross-slope walking should be utilized whenever possible to reduce the risks and limit energy consumption. It also shows that the planar feet' upsides (non-excessive sinkage) outweigh its downsides (exceeding slip and energy consumption in highly inclined terrain) compared to the point feet in this mission scenario.

\begin{table}[!tbh]
\caption{Quantitative results from the global path planner for the descent and the ascent with the planar and point feet. The histograms show the slope distributions along the robot x-axis along the path that the robot would encounter. For all energy-based quantities, we provide estimates for martian values for comparison (Mars' gravity is one third of Earth's gravity).}
\centering
\begin{threeparttable}
\begin{tabular}{
| l |  S[table-format=4.2, table-column-width=1.50cm] l | S[table-format=4.2, table-column-width=1.50cm]  l |  S[table-format=4.2, table-column-width=1.50cm]  l | S[table-format=4.2, table-column-width=1.50cm]  l |
} 
\hline
    Direction                & \multicolumn{4}{c|}{Ascent} &  \multicolumn{4}{c|}{Descent} \\ \cline{1-9}
    Foot type                & \multicolumn{2}{c|}{Point foot}    & \multicolumn{2}{c|}{Planar foot}       & \multicolumn{2}{c|}{Point foot}     & \multicolumn{2}{c|}{Planar foot}    \\ \cline{1-9}

                    & \multicolumn{2}{c|}{\includegraphics[width=0.19\textwidth]{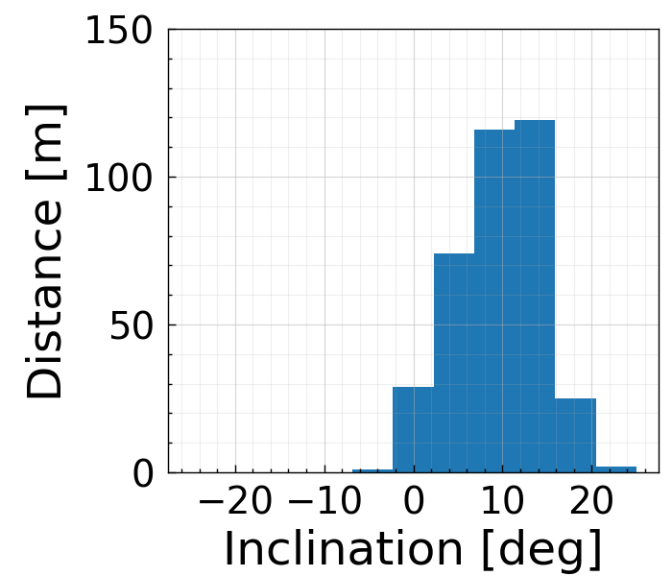}} 
                    & \multicolumn{2}{c|}{\includegraphics[width=0.19\textwidth]{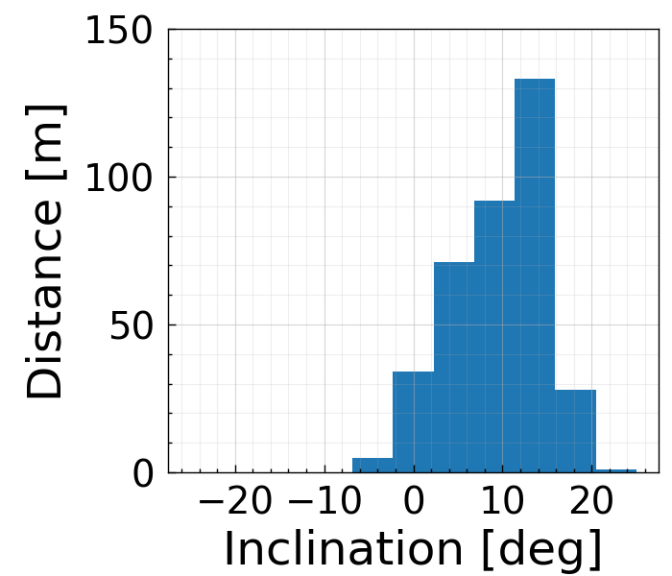}} 
                    & \multicolumn{2}{c|}{\rule{0pt}{80pt}\includegraphics[width=0.19\textwidth]{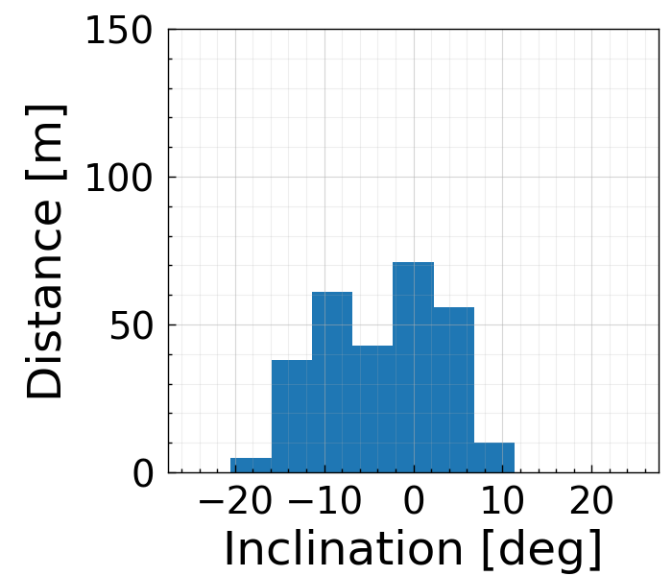}} 
                    & \multicolumn{2}{c|}{\includegraphics[width=0.19\textwidth]{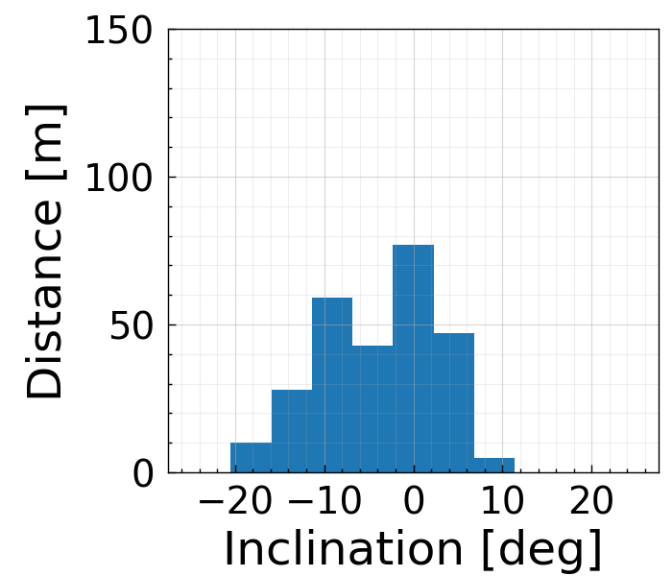}}\\ \hline
Distance & \SI{366}{m} &  & \SI{364}{m} &  & \SI{284}{m}  &  &  \SI{269}{m} & \\ \hline 
Energy & \SI{340}{kJ} & (\SI{113}{kJ}**) & \SI{368}{kJ} & (\SI{123}{kJ}**) & \SI{364}{kJ} & (\SI{88}{kJ}**) & \SI{234}{kJ} & (\SI{78}{kJ}**)  \\ \hline 
Time & \SI{1299}{s} & & \SI{1346}{s} & & \SI{1116}{s} & & \SI{1042}{s} & \\ \hline 
Power* & \SI{262}{W} &(\SI{87}{W}**)& \SI{274}{W} &(\SI{91}{W}**)&\SI{237}{W}&(\SI{79}{W}**)& \SI{225}{W} & (\SI{75}{W}**) \\ \hline 
Norm. Energy*  &\SI{929}{J/m}&(\SI{310}{J/m}**)&\SI{1012}{J/m}&(\SI{337}{J/m}**)&\SI{930}{J/m}& (\SI{310}{J/m}**)&\SI{868}{J/m}&(\SI{290}{J/m}**)\\ \hline 
Velocity* &\SI{0.28}{m/s}& &\SI{0.27}{m/s}& &\SI{0.25}{m/s}& &\SI{0.26}{m/s}& \\ \hline 

\end{tabular}
 \begin{tablenotes}
      \small
    \item  * Power, velocity and J/m are average values over entire path length.
    \item  ** Approximated martian values (divided by 3.0 to approximate martian gravity conditions) 
    \end{tablenotes}
    \end{threeparttable}
\label{tab:spacebok_path_planner_quantitative_results}
\end{table}

\section{Discussion}\label{sec:slope_climbing_discussion}

\subsection{Limitations of single-foot experiments}
Due to our testbed's technical limitations, all single-foot traction tests were carried out on planar soil. However, on dry, granular material, traction is also controlled by the inclination of the slope, as the soil tends to shear easily when being close to its angle of internal friction. To adequately test this phenomenon, additional tests at various inclinations with a broad set of simulants would be required. Similarly, we only used one specific load during testing: While this load is representative of the weight of our robot, more experiments are required to establish a general, experimental sinkage/traction model for weight, foot design, slope, and soil properties. Possible future investigations should also aim at gaining a better theoretical (and practical) understanding of the grouser-soil interaction for legged systems. While the grousers clearly provide an advantage, understanding the underlying principles that govern the design would help find an optimal length, number, shape, and spacing.

\subsection{Foot performance and safety considerations}
We found it difficult to clearly define "safety" for the system. For example, the point feet provide a relatively stable stance and sufficient traction on steep slopes, as they experience excessive sinkage, ultimately locking the foot in the soil. Consequently, most of the foot trajectory is below the surface, which increases the risk of falls if unperceived obstacles are encountered. In this context, NASA's \textit{Spirit} rover got immobilized after breaking through a solid-looking, crusty surface, highlighting the risks of unknown subsurface composition \cite{webster2009nasa}. The sinkage also resulted in motions close to the leg's singularity configuration when walking in parallel to the slope. This would render the traversal of softer soils, like ES-1 or TUBS, impossible. On the other hand, the planar feet generally performed well, but slippage increased significantly on slopes close to the soil's angle of response. This is due to the limited sinkage of the planar feet into the soil, which allowed the top layers of simulant to shear off when applying a horizontal force due to a lack of cohesion (see, e.g., \cite{terzaghi, bickelandkring}).

Additionally, the relatively large surface of the planar feet also accumulated sand over time on steep inclinations. For a future iteration, the top plate and connection to the leg should be designed conically to minimize the accumulation of sand. The surface area of the planar feet should also be reviewed carefully. If worst-case scenarios, such as ES-1 type soils, could be avoided, feet with a smaller surface area could be considered, as they allow for an optimization of the trade-off between sinkage and traction, reducing the risk of falling while maximizing mobility. Such feet would then allow sinkage to an acceptable level which reduces the risk of falling while maximizing traction.

\subsection{Locomotion control and gait selection}
In the scope of our experiments, we used a static and dynamic gait with hand-tuned gait parameters for a representative, although specific, soil. While these worked well and our tunings represent a good starting point, they might not generalize to other soils. To overcome this issue, a library of several soil-dependant gaits and parameters could be constructed, where the robot selects the appropriate gait based on proprioceptive terrain classification \cite{kolvenbach2019ral}, for example. For extremely steep terrains, even more, exotic gaits such as inching might be considered \cite{remy2010alof}. Alternatively, or in addition, a more reactive approach can be taken, where the robot only adapts its parameters based on the occurrence of certain phenomena, such as slippage \cite{jenelten2019slippery}.  A new yet promising approach is to learn locomotion in simulation with deep reinforcement learning (RL) \cite{hwangbo2019learning}. Training a walking policy on a broad set of grounds might present a solution to increase overall robustness, as shown in recent work on natural terrain \cite{Lee2020science}. The downside of such approaches is the extensive tuning of cost functions and careful crafting of the simulation environment to minimize the reality gap. Additionally, due to the approach's non-deterministic nature, RL policies for space might require extensive validation tests to guarantee a statistical success ratio, which in turn drives cost.

\subsection{Actuation concept limitations}

Our robot uses only two DOF per leg. While we could perform all foreseen experiments and gain from the favorable power-to-weight ratio and reduced complexity, an additional hip abduction/deduction DOF would undoubtedly increase the system's robustness and overall mobility. For example, evasive steps could help regain balance after perturbations, and (point-) turns without violating friction constraints would allow for higher stability and more versatile path-planning. The hip DOF would also enable the ability of the robot to recover from a fall, as we have demonstrated on \textit{ANYmal} \cite{hwangbo2019learning}.

A significant portion of our robot's energy consumption is attributed to the suspension of the body weight. This high-torque, low-speed regime is particularly inefficient for rotary Brushless DC motors \cite{mahmoudi2015efficiency}. Since the generated motor torque $\tau_m$ is proportional to the current $I_m$, $\tau_m \propto I_m $ and thermal losses $P_j$ are proportional to the current squared $ P_j \propto I_m^2 $. As a result, the motors start to heat up significantly during the stance. Excess heat and inefficient operation are two main concerns that have to be addressed for deployment in space. Even though the gravitational acceleration on Mars is only a third compared to Earth, a passive suspension system, such as parallel springs or a variable transmission, should be considered. 

\subsection{Extending path-planning strategies for Mars}
The sampling-based planner showed that it is possible to find safe and energy-efficient routes through martian craters. A general strategy is hereby to traverse slopes in curves while staying as close as possible to an optimal AOA. We used a relatively simple cost function in our setup, which assumes that only soils similar to ES-3 would be encountered. On Mars, however, various soils with different terramechanical properties are going to be encountered. In this context, orbital image classifiers, such as \textit{SPOC} \cite{rothrock2016spoc} show that it is possible to classify soil types on the surface of Mars ahead of a mission. Together with more data on the energy consumption and velocity of the robot on various terrains, a further refinement of the planner could be achieved.

\section{Conclusion}\label{sec:slope_climbing_conclusion}
We developed a point- and a passive-adaptive planar foot and demonstrated that a dynamic quadrupedal robot could use them to successfully traverse steep slopes of Martian soil simulant (ES-3) up to 25\degree~(the maximum angle achievable in the facility). The experimental analysis of different foot designs suggests that an adaptive planar foot surface area of at least \SI{110}{cm^2} significantly reduces sinkage on inclined and highly compressible soils (such as ES-1) and that long grouser of around \SI{12}{mm} boost traction on soft and stiff lunar and martian regolith simulants. We additionally observe that a pointy grouser design increases the chance of interlocking with rigid, natural surfaces such as bedrock. We note that point feet can provide better traction than adaptive planar feet on slopes that approach the material's angle of internal friction due to their increased sinkage - but point out that the potential for excessive instability and tripping is increased as well. If the absence of highly compressible soils can be guaranteed, planar feet with diameters smaller \SI{110}{cm^2} would present a valid alternative.

We present a torque-controlled locomotion controller that allows the robot to estimate the inclination of the slope and adapt the torso position such that its feet are loaded equally. Two walking patterns, a static walk and a dynamic trotting motion are experimentally tuned to allow for safe and efficient walking on the slope, with varying AOA. We further compare point and planar feet performance and the static and dynamic gait in terms of energy consumption and velocity and find that below \SI{20}{\degree} all combinations of feet and gait patterns allow for a safe traverse. A dynamic trotting motion is shown to significantly reduce the energy consumption per meter, mainly because it allows for walking at twice the speed of a static walk and reduces the contribution of energy required to suspend the robot's weight, which represents a dominant fraction of the energy budget under earth gravity.

Lastly, we utilize the derived energy information to plan hypothetical, energy-efficient paths through an exemplary martian crater using a custom RRT* implementation. Our results show that diagonal walks, limiting the inclination in heading direction below 20\degree, are more energy-efficient (and potentially safer) than straight,  head-on climbs. The results also reveal only marginal differences in terms of energy between both foot designs.

While our general, experimental approach highlights the potential of dynamic quadrupeds for future planetary exploration missions, further research and developments are necessary to increase the robustness and reliability. Firstly, end-to-end locomotion testing was only performed on a single regolith simulant. To evaluate the robot's performance more thoroughly, additional tests on different soils and rocky terrains are necessary. Generally, the physical interplay between feet, soil, slope, and walking controller is not well understood, and more empirical and theoretical work is required to establish optimal foot and controller designs that also generalize well to different environments.

In terms of hardware, a hip abduction/deduction DOF on the robot would significantly increase stability, especially during on-slope turning. This, in turn, would also allow for a more sophisticated path planner, which could make use of the increased mobility of the robot. Passively compensating for the weight of the robot would further reduce energy consumption. A limitation of the planar foot is the risk of soil accumulation on the top plate and twisting/jamming in front of small rocks, which could be addressed in future design iterations.

\subsubsection*{Acknowledgments} The authors would like to thank Stefan Linke and the Institute of Space Systems at the Technical University of Braunschweig for providing access to lunar \textit{TUBS} simulant and Aaron Jordan for the rewarding discussions about establishing and maintaining uniform soil conditions during the testing campaign. This work would not have been possible without the generous offer to perform testing at RUAG Space and the support from Philipp Oettershagen and his team. This work has been supported by the European Space Agency (ESA) and Airbus DS in the framework of the Network Partnering Initiative 481-2016 and the Swiss Space Center as part of the Call for Ideas (CfI) 2019.

\bibliographystyle{apalike}
\bibliography{bibtex}

\end{document}

%% file: Figures/control/03_staticwalk.tikz
\usetikzlibrary{shapes.geometric, arrows}
\tikzstyle{arrow1} = [thick,->,>=stealth]
\tikzstyle{arrow2} = [thick,<->,>=stealth]

\definecolor{color1}{rgb}{0.8431, 0.1882, 0.1216}
\definecolor{color2}{rgb}{0.9882, 0.5529, 0.3490}
\definecolor{color3}{rgb}{0.9922, 0.8000, 0.5412}
\definecolor{color4}{rgb}{0.9961, 0.9412, 0.8510}


\begin{tikzpicture}[scale=0.45]
\filldraw[fill=color2, draw=black] (2,0) rectangle (15,1);
\filldraw[fill=color4, draw=black] (2+13*0.05, 0) rectangle (2+13*0.25,1);
\filldraw[fill=color2, draw=black] (2,1) rectangle (15,2);
\filldraw[fill=color4, draw=black] (2+13*0.3, 1) rectangle (2+13*0.5,2);
\filldraw[fill=color2, draw=black] (2,2) rectangle (15,3);
\filldraw[fill=color4, draw=black] (2+13*0.55, 2) rectangle (2+13*0.75,3);
\filldraw[fill=color2, draw=black] (2,3) rectangle (15,4);
\filldraw[fill=color4, draw=black] (2+13*0.8, 3) rectangle (2+13*1,4);
\draw[](0,0) rectangle (2,1) node[pos=.5] {RH};
\draw[](0,1) rectangle (2,2) node[pos=.5] {RF};
\draw[](0,2) rectangle (2,3) node[pos=.5] {LH};
\draw[](0,3) rectangle (2,4) node[pos=.5] {LF};

\draw[arrow1] (2+13*0.05,-0.5) -- (2+13*0.05,0);
\draw[arrow1] (2+13*0.25,-0.5) -- (2+13*0.25,0);

\draw[] (2,0) -- (2,-2);
\draw[] (15,0) -- (15,-2);

\draw[draw=white](2+13*0.05, -0.5) rectangle (2+13*0.05,-1) node[pos=.5] {TO};
\draw[draw=white](2+13*0.25, -0.5) rectangle (2+13*0.25,-1) node[pos=.5] {TD};
\draw[draw=white](17/2, -1.2) rectangle (17/2,-2) node[pos=.5] {Gait cycle time};
\draw[arrow2] (2,-2) -- (15,-2);

\end{tikzpicture}


%% file: Figures/control/04_trot.tikz
\usetikzlibrary{shapes.geometric, arrows}
\tikzstyle{arrow1} = [thick,->,>=stealth]
\tikzstyle{arrow2} = [thick,<->,>=stealth]

\definecolor{color1}{rgb}{0.8431, 0.1882, 0.1216}
\definecolor{color2}{rgb}{0.9882, 0.5529, 0.3490}
\definecolor{color3}{rgb}{0.9922, 0.8000, 0.5412}
\definecolor{color4}{rgb}{0.9961, 0.9412, 0.8510}


\begin{tikzpicture}[scale=0.45]
\filldraw[fill=color2, draw=black] (2,0) rectangle (15,1);
\filldraw[fill=color4, draw=black] (2+13*0.1, 0) rectangle (2+13*0.4,1);
\filldraw[fill=color2, draw=black] (2,1) rectangle (15,2);
\filldraw[fill=color4, draw=black] (2+13*0.6, 1) rectangle (2+13*0.9,2);
\filldraw[fill=color2, draw=black] (2,2) rectangle (15,3);
\filldraw[fill=color4, draw=black] (2+13*0.6, 2) rectangle (2+13*0.9,3);
\filldraw[fill=color2, draw=black] (2,3) rectangle (15,4);
\filldraw[fill=color4, draw=black] (2+13*0.1, 3) rectangle (2+13*0.4,4);
\draw[](0,0) rectangle (2,1) node[pos=.5] {RH};
\draw[](0,1) rectangle (2,2) node[pos=.5] {RF};
\draw[](0,2) rectangle (2,3) node[pos=.5] {LH};
\draw[](0,3) rectangle (2,4) node[pos=.5] {LF};

\draw[arrow1] (2+13*0.1,-0.5) -- (2+13*0.1,0);
\draw[arrow1] (2+13*0.4,-0.5) -- (2+13*0.4,0);

\draw[] (2,0) -- (2,-2);
\draw[] (15,0) -- (15,-2);

\draw[draw=white](2+13*0.1, -0.5) rectangle (2+13*0.1,-1) node[pos=.5] {TO};
\draw[draw=white](2+13*0.4, -0.5) rectangle (2+13*0.4,-1) node[pos=.5] {TD};
\draw[draw=white](17/2, -1.2) rectangle (17/2,-2) node[pos=.5] {Gait cycle time};
\draw[arrow2] (2,-2) -- (15,-2);

\end{tikzpicture}
